\documentclass[dvipsnames]{article} %
\usepackage{colm2024_conference}

\usepackage{booktabs}
\usepackage{enumitem}
\usepackage{wrapfig}
\usepackage{algorithm}
\usepackage{algpseudocode}
\usepackage{graphicx}
\usepackage{subfigure}
\usepackage[misc]{ifsym}

\usepackage{microtype}
\usepackage{amsmath}
\usepackage{colortbl}
\usepackage[utf8]{inputenc}
\usepackage[T1]{fontenc}
\definecolor{lightgray}{rgb}{0.9,0.9,0.9}
\usepackage{caption}
\usepackage{subcaption}
\usepackage{setspace}
\usepackage{url}
\usepackage{multirow}
\usepackage{tabularx}
\usepackage{blindtext}
\usepackage{pgfplots}
\pgfplotsset{compat=1.18} 
\usepackage{tikz}
\usetikzlibrary{er,positioning,bayesnet}
\usepackage{makecell}
\usepackage{tipa}
\usepackage{siunitx}
\usepackage{nicefrac}
\usepackage{listings}
\usepackage[raster,skins, most]{tcolorbox} %
\usepackage{xltabular}
\usepackage{adjustbox}
\usepackage{xurl}
\usepackage{rotating}
\usepackage[normalem]{ulem}

\useunder{\uline}{\ul}{}

\usepackage{amsmath,amsfonts,bm}

\def\eqref#1{equation~\ref{#1}}

\def\1{\bm{1}}

\DeclareMathAlphabet{\mathsfit}{\encodingdefault}{\sfdefault}{m}{sl}
\SetMathAlphabet{\mathsfit}{bold}{\encodingdefault}{\sfdefault}{bx}{n}

\newcommand*\justify{%
  \fontdimen2\font=0.4em
  \fontdimen3\font=0.2em
  \fontdimen4\font=0.1em
  \fontdimen7\font=0.1em
  \hyphenchar\font=`\-
}

\renewcommand{\texttt}[1]{%
  \begingroup
  \ttfamily
  \begingroup\lccode`~=`/\lowercase{\endgroup\def~}{/\discretionary{}{}{}}%
  \begingroup\lccode`~=`[\lowercase{\endgroup\def~}{[\discretionary{}{}{}}%
  \begingroup\lccode`~=`.\lowercase{\endgroup\def~}{.\discretionary{}{}{}}%
  \catcode`/=\active\catcode`[=\active\catcode`.=\active
  \justify\scantokens{#1\noexpand}%
  \endgroup
}

\usepackage{makecell}
\usetikzlibrary{tikzmark}
\makeatletter
\newcommand*\myfontsize{%
  \@setfontsize\myfontsize{7}{8}%
}
\makeatother

\definecolor{uclablue}{RGB}{159, 195, 224}

\definecolor{uclagold}{RGB}{255, 240, 180}

\definecolor{aliceblue}{RGB}{255, 238, 241}

\definecolor{cadmiumgreen}{rgb}{0.0, 0.42, 0.24}

\definecolor{myred}{rgb}{0.7, 0.3, 0.0}
\definecolor{myblue}{rgb}{0.2, 0.3, 0.6}
\definecolor{babygreen}{rgb}{0.85, 0.97, 0.85}

\definecolor{purple1}{RGB}{126, 107, 196}
\definecolor{purple2}{RGB}{199, 158, 207}
\definecolor{purple3}{RGB}{214, 200, 255}
\definecolor{purple4}{RGB}{254, 240, 255}

\definecolor{deepblue}{RGB}{48, 58, 82}

\definecolor{deepPurple}{HTML}{330066}

\definecolor{uclablue_old}{rgb}{0.15, 0.45, 0.68}
\hypersetup{
    breaklinks,
    citecolor=uclablue_old,
    colorlinks=true,
}

\newtcolorbox{mybox}[2][]
  {colback = black!5!white, colframe = black!75!black, fonttitle = \bfseries,
    colbacktitle = black!100!black, enhanced, before upper={\fontsize{8}{11}\obeyspaces\obeylines\selectfont}, fontupper=\selectfont,
    attach boxed title to top left={yshift=-2.2mm,xshift=4mm},
    title=#2,#1}

\title{Fun-Audio-Chat Technical Report}

\author{Tongyi Fun Team, Alibaba Group}

\newcommand{\huggingface}{\raisebox{-1.5pt}{\includegraphics[height=1.05em]{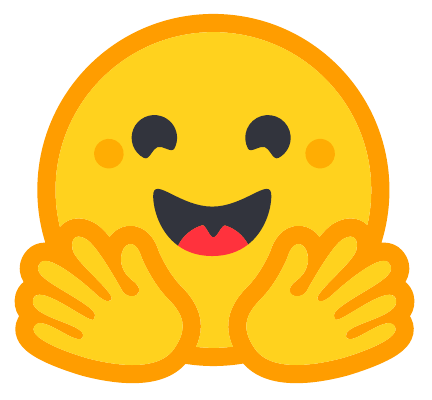}}\xspace}
\newcommand{\modelscope}{\raisebox{-1.5pt}{\includegraphics[height=1.05em]{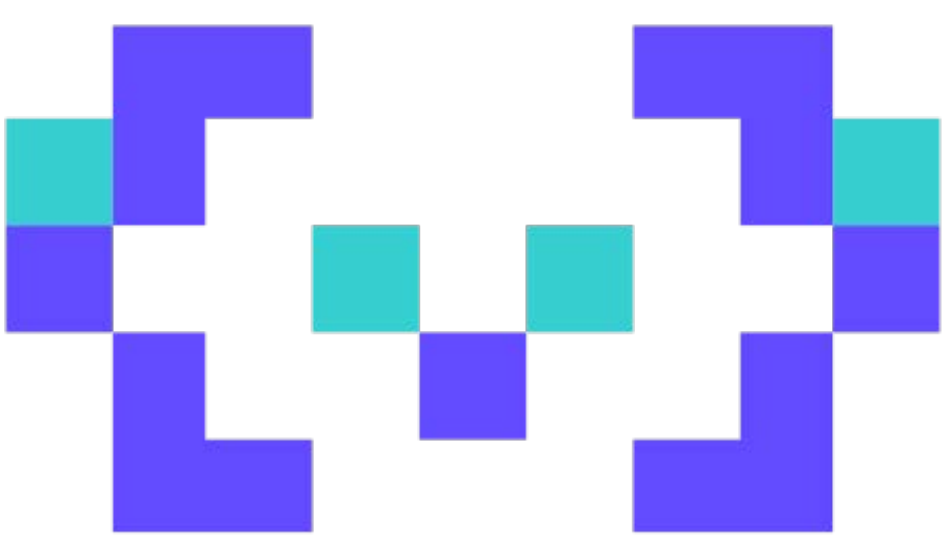}}\xspace}
\newcommand{\github}{\raisebox{-1.5pt}{\includegraphics[height=1.05em]{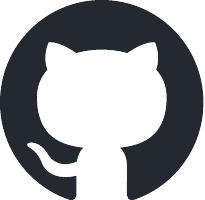}}\xspace}
\newcommand{\worldwideweb}{\raisebox{-1.5pt}{\includegraphics[height=1.05em]{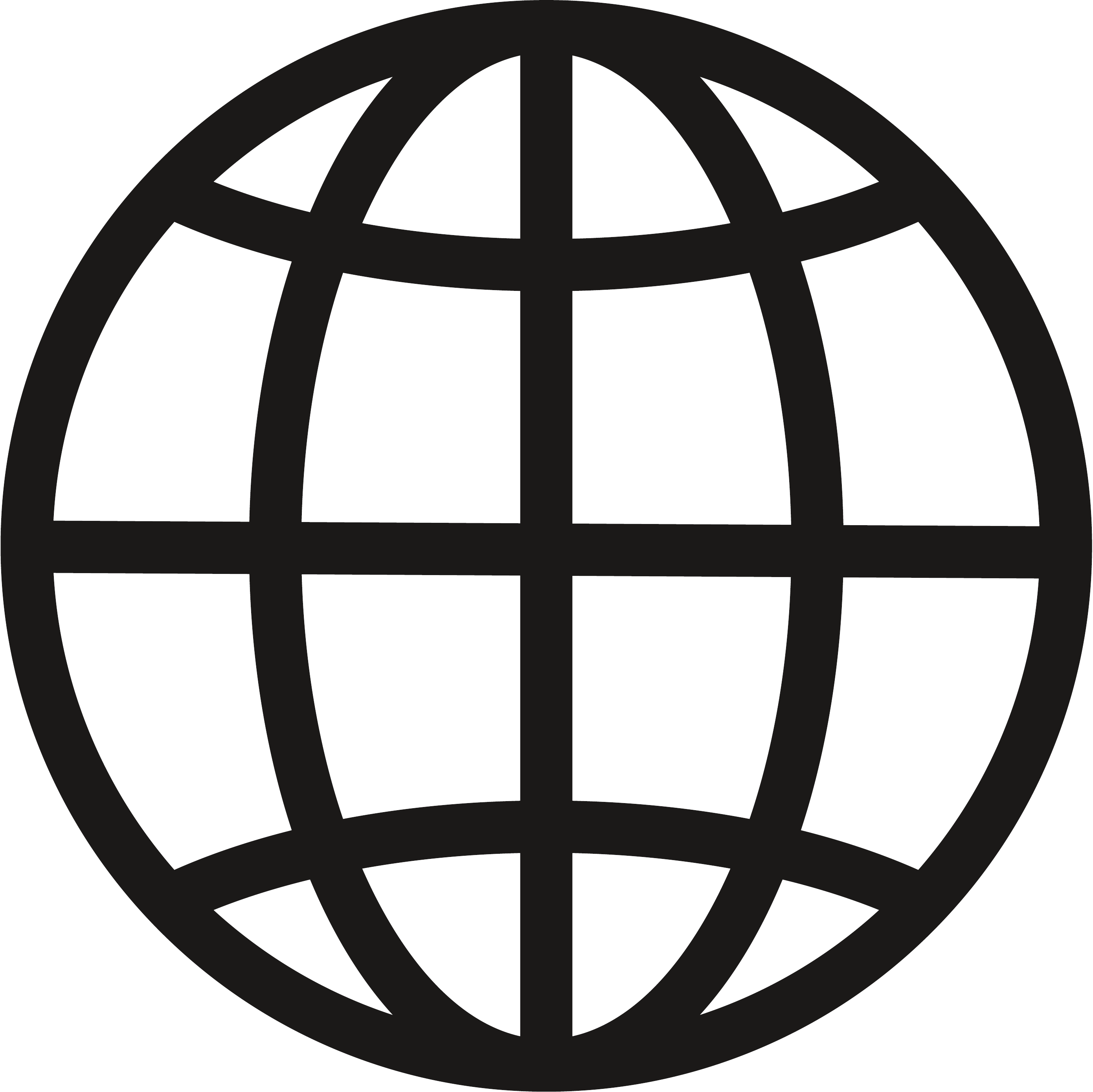}}\xspace}

\begin{document}

\maketitle

\begingroup
  \renewcommand\thefootnote{\Letter}  
  \footnotetext{FunAudioLLM@list.alibaba-inc.com} 
\endgroup

\vspace{-20pt}

\begin{abstract}
Recent advancements in joint speech-text models have demonstrated great potential for seamless voice interactions. However, existing models face critical challenges: the temporal resolution mismatch between speech tokens (typically 25Hz) and text tokens (approximately 3Hz) dilutes semantic information, incurs high computational costs limit practical deployment, and leads to catastrophic forgetting of text LLM knowledge during multimodal training. In this work, we introduce \textbf{Fun-Audio-Chat}, a Large Audio Language Model (LALM) that addresses these limitations by adopting two key innovations from our previous work DrVoice.  First, we employ the \textbf{Dual-Resolution Speech Representations (DRSR)} architecture: the Shared LLM backbone processes audio at an efficient 5Hz frame rate (achieved through speech token grouping), while the Speech Refined Head (SRH) generates high-quality speech tokens at 25Hz resolution. This dual-resolution design effectively balances computational efficiency (reducing GPU hours by nearly 50\%) and speech generation quality. Second, we adopt the \textbf{Core-Cocktail Training strategy} in full supervised fine-tuning, a two-stage training approach with intermediate model merging that mitigates catastrophic forgetting. After Core-Cocktail training, we introduce \textbf{Multi-Task DPO Training} to enhance robustness, audio understanding, instruction-following and voice empathy capabilities. This multi-stage post-training paradigm enables Fun-Audio-Chat to effectively retain knowledge of the original text LLM while gaining powerful audio understanding, reasoning, and generation skills.  Different from the majority of recent LALMs that rely on both large-scale audio-text pre-training and post-training to develop audio capabilities, \textbf{Fun-Audio-Chat only leverages pre-trained models and utilizes extensive post-training}. Fun-Audio-Chat dense 8B and MoE 30B-A3B models achieve competitive performance on Speech-to-Text and Speech-to-Speech generation tasks, ranking Top among models of similar scales across multiple Spoken Question Answering benchmarks. It also achieves competitive to superior performance on Audio Understanding, Speech Function Calling, Speech Instruction-Following and Voice Empathy benchmarks. We further develop \textbf{Fun-Audio-Chat-Duplex}, a full-duplex variant that achieves strong performance on Spoken Question Answering benchmarks and full-duplex interactions. We open-source the Fun-Audio-Chat-8B model checkpoint with its training and inference code, and provide an interactive demo. 

\end{abstract}

\begin{center}
    \small
    \begin{tabular}{rll}
        \github{} & \textbf{GitHub} & \url{https://github.com/FunAudioLLM/Fun-Audio-Chat} \\
        \huggingface{} & \textbf{HuggingFace} & \url{https://huggingface.co/FunAudioLLM/Fun-Audio-Chat-8B} \\
        \modelscope{} & \textbf{ModelScope} & \url{https://modelscope.cn/FunAudioLLM/Fun-Audio-Chat-8B} \\
        \worldwideweb{} & \textbf{Demo Page} & \url{https://funaudiollm.github.io/funaudiochat} \\
    \end{tabular}
\end{center}

\begin{figure}[h]
    \centering
    \begin{minipage}{0.49\textwidth}
        \centering
        \includegraphics[width=\linewidth]{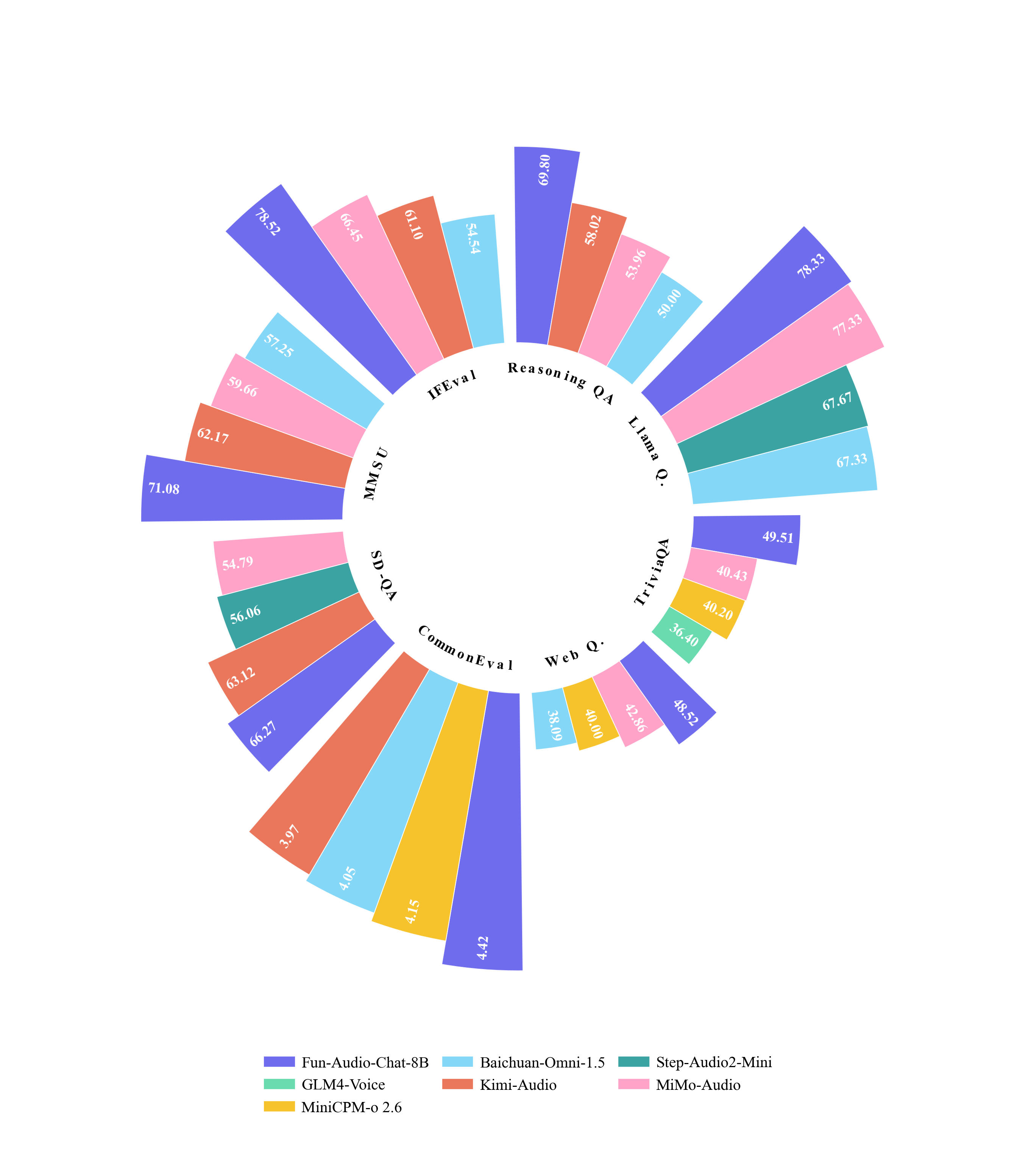}
        \caption*{(a) Performance comparison on Spoken QA tasks.}
    \end{minipage}
    \hfill
    \begin{minipage}{0.49\textwidth}
        \centering
        \includegraphics[width=\linewidth]{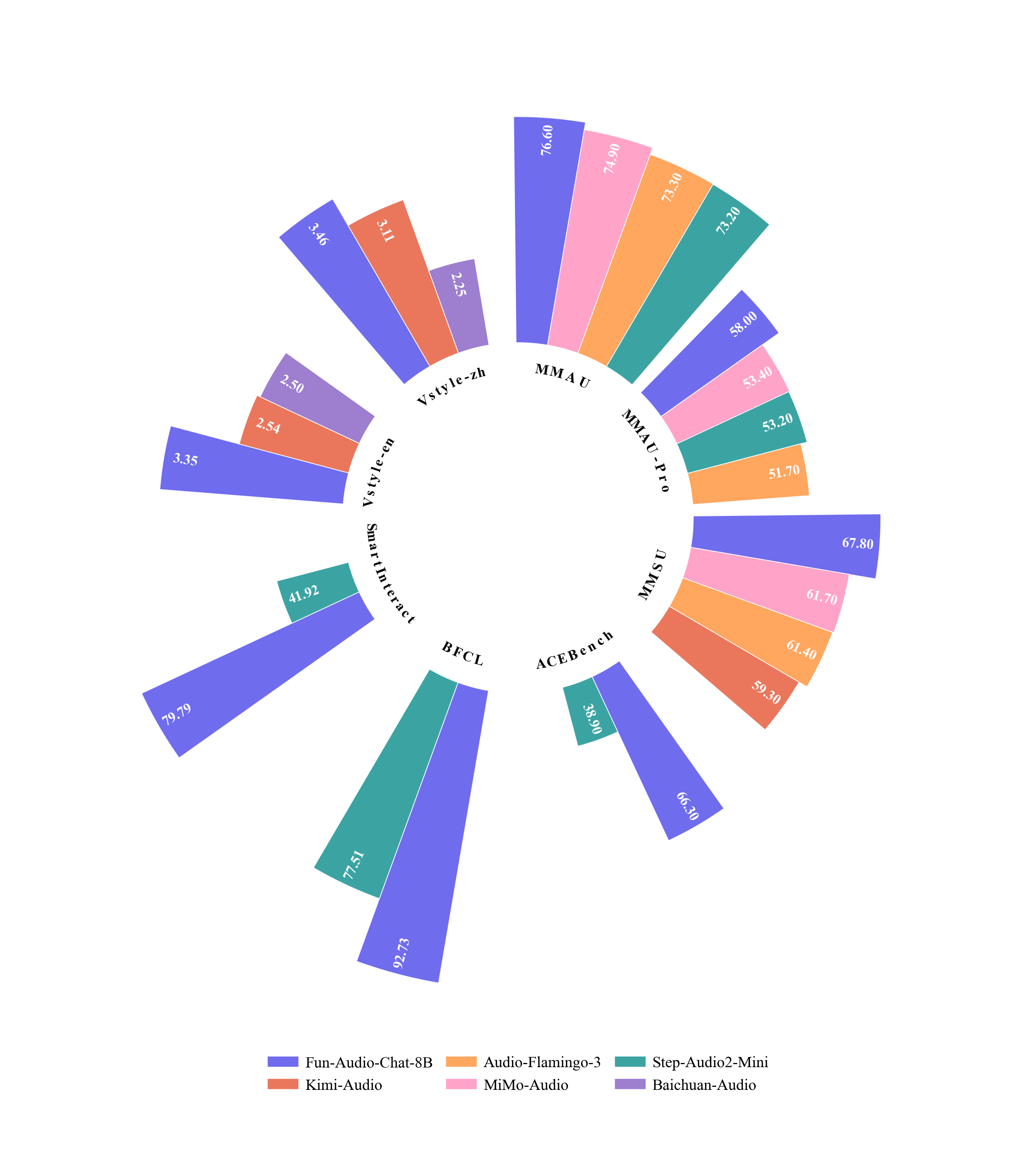}
        \caption*{(b) Performance comparison on other tasks.}
    \end{minipage}
    \caption{Performance comparison between our Fun-Audio-Chat-8B and previous $\sim$8B-scale state-of-the-art (SOTA) models across multiple benchmarks. (a) illustrates results on Spoken Question Answering benchmarks (Speech-to-Speech SQA on LlamaQ, TriviaQ, WebQ in UltraEvalAudio; Speech-to-Text SQA on ReasoningQA in OpenAudioBench; Speech-to-Text SQA on CommonEval, SD-QA, MMSU, and IFEval in VoiceBench), while (b) presents results on Audio Understanding (MMAU, MMAU-Pro, MMSU), Speech Function Calling (Speech-ACEBench,  Speech-BFCL, Speech-SmartInteract), Speech Instruction-Following and Voice Empathy (VStyle, English and Mandarin subsets) benchmarks. Detailed evaluations are presented in Section~\ref{sec:experiments}.}
    \label{fig:performance_overview}
\end{figure}

\section{Introduction}
\label{sec:introduction}

The development of spoken dialogue systems is critical to human-computer interaction, as natural human communication inherently relies on verbal exchanges. Recently, Large Language Model (LLM) based spoken dialogue systems, exemplified by systems like GPT-4o~\citep{openai2024hello}, demonstrate great potential for seamless and natural voice interactions with users. LLM-based spoken dialogue systems can be generally categorized into \textbf{cascaded} and \textbf{end-to-end (E2E)} systems, with the distinction lying in whether the backbone LLM can \textit{directly} comprehend speech representations and generate speech outputs.

Many recent E2E models focus on \textbf{Joint Speech-Text Models}~\citep{DBLP:journals/corr/abs-2410-00037, chen2024slam, kimiteam2025kimiaudiotechnicalreport}, where LLMs take speech representations as input and generate \textit{both text tokens and speech tokens simultaneously}. However, existing joint speech-text models face critical challenges: (1) the temporal resolution mismatch between speech tokens (typically 25Hz) and text tokens (approximately 3Hz)~\citep{chen2024slam} dilutes semantic information and hinders the full utilization of the LLM's core capabilities; (2) continual pre-training and post-training the text-LLM backbone into multimodal models often lead to catastrophic forgetting of the text LLM's knowledge; (3) high computational costs due to high audio frame rates (typically 12.5Hz or 25Hz), limiting practical deployment.

In this work, we present \textbf{Fun-Audio-Chat}, a parallel large audio language model (LALM) that extends our previous work DrVoice~\citep{tan2025drvoiceparallelspeechtextvoice} by adopting the \textbf{Dual-Resolution Speech Representations (DRSR)} architecture and scaling it up to significantly larger training datasets of millions of hours of diverse audio data and larger model scales (dense 8B and MoE 30B-A3B~\footnote{30B-A3B denotes a Mixture-of-Experts (MoE) model with 30B total parameters and 3B active parameters.}).  

For speech comprehension in Fun-Audio-Chat, we employ a grouping mechanism that maps 25Hz audio tokens to 5Hz speech representations, enabling the shared LLM backbone to process audio at an efficient 5Hz frame rate. During generation, the hidden states from the shared LLM layer are passed in parallel to a Text Head for text token prediction and a Speech Refined Head (SRH) to generate high-quality speech tokens at 25Hz resolution. This dual-resolution design effectively balances computational efficiency (reducing GPU hours by nearly 50\%) and speech generation quality.

The majority of recent open-source LALMs and Omni-language-models rely on both large-scale audio-text pre-training (e.g., audio/text unimodal pre-training, audio-text mapping and interleaving pre-training tasks) and post-training to develop strong audio capabilities, such as Kimi-Audio~\citep{kimiteam2025kimiaudiotechnicalreport}, Step-Audio 2~\citep{DBLP:journals/corr/abs-2507-16632}, MiMo-Audio~\citep{coreteam2025mimoaudio}, and Longcat-Flash-Omni~\citep{DBLP:journals/corr/abs-2511-00279}. In contrast, Fun-Audio-Chat leverages pre-trained models and is trained with a multi-stage post-training paradigm, without large-scale audio-text pre-training (similarly, Audio-Flamingo-3~\citep{DBLP:journals/corr/abs-2507-08128} also does not use large-scale audio-text pre-training). After initialization from text-based or vision-language LLMs, the \textbf{Pre-alignment} stage updates the audio encoder, the adapter, and the Speech Refined Head using large-scale speech-text paired data. We then adopt the \textbf{Core-Cocktail Training strategy} proposed in our earlier work DrVoice~\citep{tan2025drvoiceparallelspeechtextvoice}, to address catastrophic forgetting in multimodal training. Core-Cocktail Training is a two-stage approach that involves: (1) Stage 1: fine-tuning with high learning rate to rapidly adapt the model, (2) intermediate model merging of the Stage-1 model and the original pre-trained LLM backbone to preserve knowledge, and (3) Stage 2: fine-tuning with low learning rate for stable optimization. Following Core-Cocktail Training, we conduct \textbf{Multi-Task DPO Training} to boost the robustness to real speech data and the capabilities of speech instruction-following, audio understanding, and voice empathy. This multi-stage post-training paradigm enables Fun-Audio-Chat to retain the original text-LLM's capabilities while gaining powerful audio understanding, reasoning, and generation skills.

Our contributions can be summarized as follows:

\begin{itemize}[leftmargin=*,noitemsep]
    \item \textbf{Large-Scale Post-Training and Model Scaling.} Fun-Audio-Chat scales up the two key innovations of \textbf{Dual-Resolution Speech Representations (DRSR)} architecture and \textbf{Core-Cocktail Training strategy} in our earlier work DrVoice~\citep{tan2025drvoiceparallelspeechtextvoice} to significantly larger data scales of millions of hours of diverse audio data and larger model scales, including \textbf{dense 8B and MoE 30B-A3B parameters}. This work verifies that the two key innovations in DrVoice demonstrate excellent scalability: DRSR, with its efficient 5Hz processing for the backbone LLM and 25Hz generation head, retains high computational efficiency (\textbf{approximately 50\% reduction in training GPU hours}) at larger scales; and Core-Cocktail Training strategy, with its two-stage training using different learning rates and intermediate model merging, effectively mitigates catastrophic forgetting in both 8B and 30B-A3B models. The large-scale post-training enables Fun-Audio-Chat to achieve superior performance across multiple benchmarks while maintaining the computational efficiency advantages from the dual-resolution design.
    
    \item \textbf{Multi-Task DPO Training for Enhancing Robustness and Generalizability.} Following Core-Cocktail Training, we introduce Multi-Task DPO Training to enhance the capabilities of Fun-Audio-Chat in multiple dimensions: robustness to real speech data, capabilities of instruction-following, audio understanding, and voice empathy. This training approach enables the model to better align with human preferences and improve performance on real-world conversational scenarios, distinguishing Fun-Audio-Chat from previous works that primarily rely on supervised fine-tuning. Through Multi-Task DPO training, Fun-Audio-Chat acquires advanced capabilities beyond basic speech-text interaction, including speech function calling, speech instruction-following, and voice empathy (recognizing and reasoning over user's emotional states and generating empathetic responses), enabling the model to understand and respond to complex voice interactions with appropriate emotional intelligence and functional execution.    

    \item \textbf{Comprehensive Evaluation and Strong Performance.} Extensive evaluations demonstrate that Fun-Audio-Chat 8B and 30B-A3B achieve superior performance on Spoken Question Answering (on both Speech-to-Text and Speech-to-Speech generation tasks), ranking top among models of similar scales. It also demonstrates competitive capabilities of Audio Understanding, Speech Function Calling, Speech Instruction-Following, and Voice Empathy, as demonstrated across a wide variety of commonly used benchmarks including OpenAudioBench~\footnote{\url{https://huggingface.co/datasets/baichuan-inc/OpenAudioBench}}, VoiceBench~\citep{DBLP:journals/corr/abs-2410-17196}, UltraEval-Audio~\footnote{\url{https://github.com/OpenBMB/UltraEval-Audio}}, MMAU~\citep{DBLP:conf/iclr/SakshiTKSSNDGM25}, MMAU-Pro~\citep{DBLP:journals/corr/abs-2508-13992}, MMSU~\citep{DBLP:journals/corr/abs-2506-04779}, multiple speech function calling benchmarks, and VStyle~\citep{DBLP:journals/corr/abs-2509-09716}. Detailed evaluation results are presented in Section~\ref{sec:experiments}.

    \item \textbf{Full-Duplex Voice Interaction.} We extend Fun-Audio-Chat to a full-duplex variant, \textbf{Fun-Audio-Chat-Duplex}, which supports simultaneous two-way communications. This model achieves competitive performance on Spoken Question Answering benchmarks, suggesting excellent intelligence, and strong performance in full-duplex interaction metrics (Section~\ref{sec:experiments}), demonstrating superior capabilities in natural conversation and turn-taking.

    \item \textbf{Open-source Contribution and Interactive Demo.} To promote research in this field, we open-source the dense Fun-Audio-Chat-8B model, making the model checkpoint and the training and inference code publicly available so that researchers can build upon our work. Additionally, we provide an interactive demo that showcases Fun-Audio-Chat's voice conversation capabilities.
\end{itemize}

\section{Methodology}
\label{sec:approach}

\begin{figure}[t]
    \centering
    \begin{minipage}{0.48\textwidth}
        \centering
        \includegraphics[width=\linewidth]{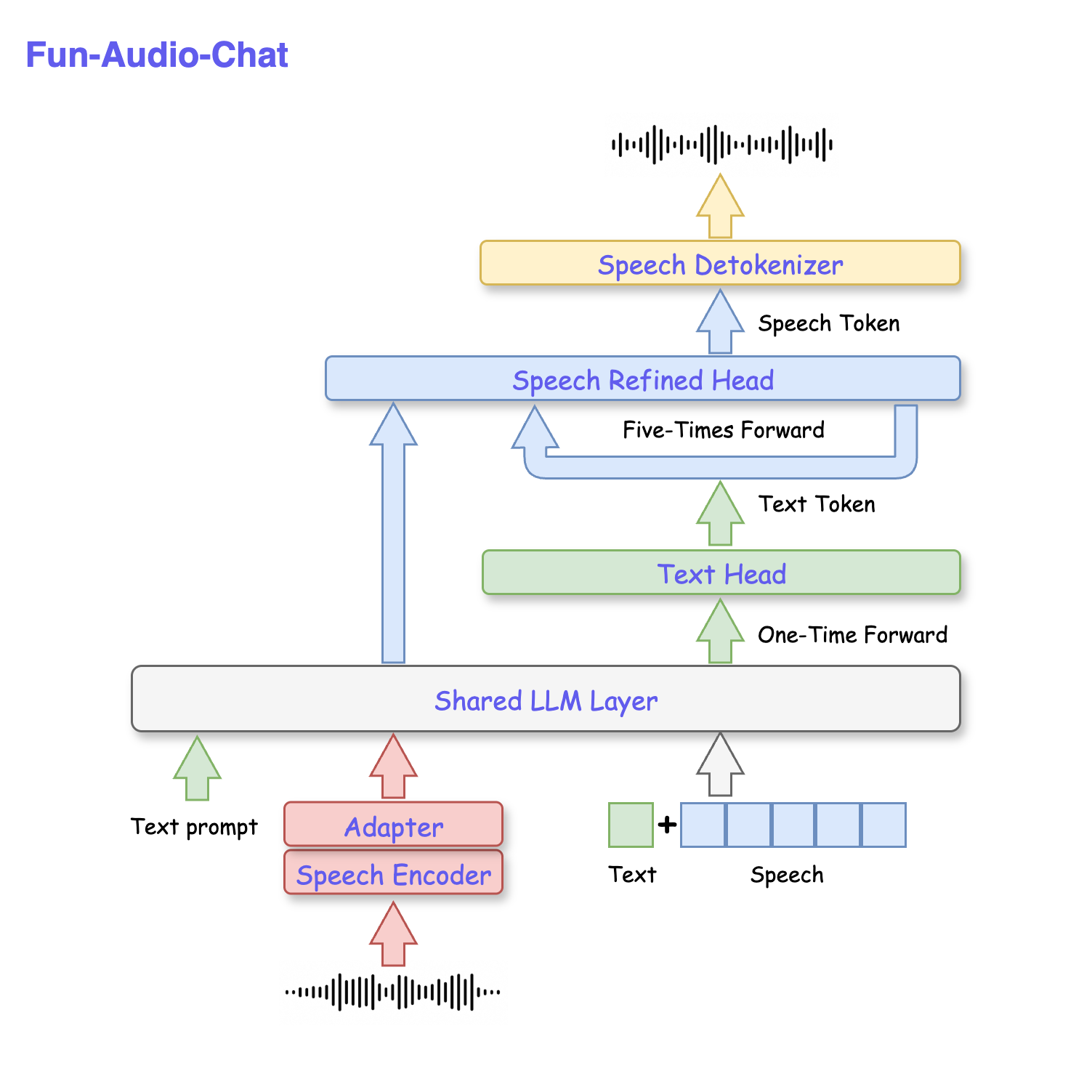}
        \caption*{(a) Fun-Audio-Chat architecture.}
    \end{minipage}
    \hfill
    \begin{minipage}{0.48\textwidth}
        \centering
        \includegraphics[width=\linewidth]{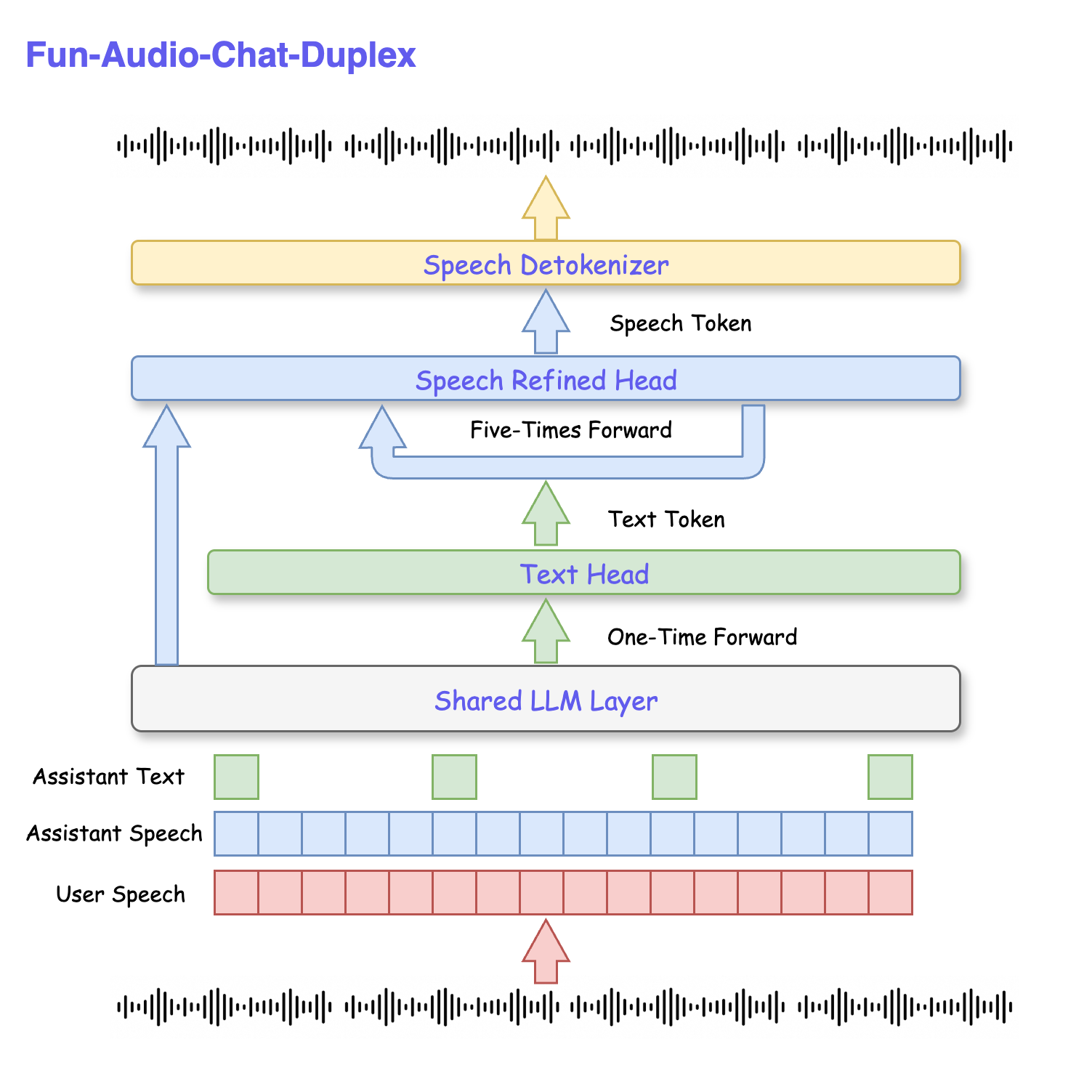}
        \caption*{(b) Full-duplex mode (Fun-Audio-Chat-Duplex).}
    \end{minipage}
    \caption{Overview of Fun-Audio-Chat. (a) User speech inputs are tokenized, \textit{grouped}, and encoded by the MLLM for autoregressive text token prediction by a \textbf{Text Head} and speech token prediction by a \textbf{Speech Refined Head (SRH)}. The MLLM comprises the \textbf{Shared LLM Layer}, the Text Head, and SRH. The generated speech tokens are then converted to speech waveform by the speech detokenizer. Note that SRH generates $5$ speech tokens through $5$ autoregressive forward passes, where $5$ is the grouping factor. (b) Full-duplex communication mode of Fun-Audio-Chat.}
    \label{fig:Fun-Audio-Chat}
  \end{figure}

Figure~\ref{fig:Fun-Audio-Chat} provides an architectural overview of Fun-Audio-Chat and its full-duplex variant Fun-Audio-Chat-Duplex. The framework of Fun-Audio-Chat comprises three primary modules: (1) For audio inputs, Speech Encoder and Speech Tokenizer transform raw audio waveforms into structured representations for both User and Assistant sides; (2) a Multimodal Large Language Model (MLLM) integrates a Shared LLM backbone with specialized Text Head and Speech Refined Head (SRH) components for generating tokens; and (3) the Speech Detokenizer reconstructs audio waveforms from the generated speech tokens. The architecture facilitates unified audio-text encoding and synchronized speech-text generation. At inference time, either text or audio inputs are converted into a common semantic representation space, which the MLLM processes to simultaneously generate both speech and text outputs via the SRH and the Text Head.

\subsection{Speech Tokenization and Detokenization}
\label{subsec:speech_tokenization}

To achieve robust audio comprehension, Fun-Audio-Chat employs Whisper-Large-v3~\citep{radford2022whisper} as the \textbf{Speech Encoder} to derive continuous representations from user audio inputs. An \textbf{Adapter} module is then applied to reduce the temporal resolution of these features and match their dimensionality to the LLM's hidden space. 
Given the demonstrated effectiveness of semantic tokens for speech representations~\citep{DBLP:conf/emnlp/ZhangLZZWZQ23, DBLP:journals/taslp/BorsosMVKPSRTGTZ23}, particularly their strong correspondence with textual content~\citep{DBLP:journals/corr/abs-2308-16692}, 
we adopt S3Tokenizer~\citep{DBLP:journals/corr/abs-2407-05407,DBLP:journals/corr/abs-2412-10117,DBLP:journals/corr/abs-2505-17589} as the \textbf{Speech Tokenizer}
to transform audio waveforms into discrete semantic token sequences $\mathbf{S}=[s_0, s_1, \cdots, s_{T-1}]$ (T denotes the sequence length) for the assistant's output.
In the reverse process, the Speech Detokenizer leverages speaker-specific embeddings that encode acoustic characteristics like timbre. The Flow Matching model~\citep{DBLP:conf/iclr/LipmanCBNL23} generates Mel-spectrogram representations from these tokens, which are then converted back to audio waveforms using the HiFi-GAN vocoder~\citep{DBLP:conf/nips/KongKB20}.

\subsection{Dual-Resolution Speech Representations (DRSR)}
\label{subsec:DRSR}
To maintain the text capabilities of pretrained text LLMs while supporting cross-modal functionality, Fun-Audio-Chat adopts the \textbf{Dual-Resolution Speech Representations (DRSR)} architecture from our earlier work DrVoice~\citep{tan2025drvoiceparallelspeechtextvoice}. This architecture effectively addresses the temporal resolution mismatch between speech tokens (typically 25Hz) and text tokens (approximately 3Hz), improves computational efficiency, and achieves high-quality speech generation.

\textbf{Speech Token Grouping.} To bridge the temporal resolution discrepancy, we apply a \textbf{grouping} technique from DrVoice~\citep{tan2025drvoiceparallelspeechtextvoice} that reduces 25Hz speech tokens to 5Hz representations for the Shared LLM backbone. The grouping transformation is expressed as follows:
\begin{equation}
    \mathbf{g}_i = \text{Linear}\left(\text{Concat}_{j=ik}^{(i+1)k-1} (\mathbf{s}_j)\right) \in \mathbb{R}^{d_{\text{text}}}
\end{equation}
where $\mathbf{s}_j$ represents individual speech tokens, $\text{Concat}$ indicates concatenation, and $k=5$ is the grouping factor based on the ratio of speech token frequency (25Hz) to the desired LLM processing frequency (5Hz). This mechanism reduces sequence length from $T$ to $T/k$, allowing the Shared LLM to operate at a 5Hz frame rate, which substantially reduces computational overhead (yielding approximately 50\% reduction in training GPU hours) while retaining the semantic reasoning abilities of the LLM.

\textbf{Speech Refined Head (SRH).} Although grouping facilitates efficient processing, it sacrifices fine-grained acoustic information essential for natural speech synthesis. To compensate this limitation, Fun-Audio-Chat integrates a specialized \textbf{Speech Refined Head (SRH)} that generates speech tokens at the complete 25Hz resolution. The SRH executes an \textbf{ungrouping} operation: the final hidden state from the Shared LLM, $\mathbf{h}_L^{\text{[SLLM]}}$, is initially transformed into group-sized embeddings through linear projection:
\begin{equation}
\mathbf{h}_{ug} = \mathbf{W}_p \mathbf{h}_L^{\text{[SLLM]}} \quad \text{where} \quad \mathbf{W}_p \in \mathbb{R}^{d_g \times d_h},
\end{equation}
which is followed by decomposition into $k$ segments:
\begin{equation}
\mathbf{H} = \text{Split}_k(\mathbf{h}_{ug}) = [\mathbf{h}_{ug}^{(1)}, \mathbf{h}_{ug}^{(2)}, \ldots, \mathbf{h}_{ug}^{(k)}],
\end{equation}
where $\mathbf{h}_{ug}^{(i)} \in \mathbb{R}^{d_{ug}/k}$. The resulting $\mathbf{H}$ provides conditional context for SRH, which generates speech tokens autoregressively at 25Hz. The training objective optimizes speech token prediction:
\begin{equation}  
    \mathcal{L}_{\text{SRH}} = -\sum_{i=1}^{T} \log P(s_i | s_{<i}, \mathbf{H}_{<i}),  
\end{equation}  
where $s_i$ denotes the $i$-th speech token. This dual-resolution framework allows Fun-Audio-Chat to simultaneously achieve computational efficiency (5Hz processing in the Shared LLM Layer) and high-fidelity speech synthesis (25Hz generation through SRH), following the design principles established in DrVoice~\citep{tan2025drvoiceparallelspeechtextvoice}.

\subsection{Multimodal Large Language Model (MLLM)}
\label{subsec:mllm}

The MLLM architecture extends pretrained text-LLMs to support \textit{unified audio-text processing}, enabling the model to handle either speech or text inputs and generate \textit{simultaneous} speech and text outputs.

Fun-Audio-Chat is a \textbf{Parallel Joint Speech-Text Model}.
Following the approach in Moshi~\citep{kyutai2024moshi}, we integrate explicit text streams to provide semantic guidance for speech generation. Our design concentrates modality alignment \textbf{solely on the assistant side}, reflecting the inherent \textbf{asymmetry} in human-computer dialogue: \textit{Users typically provide single-modality inputs (text or speech), while assistants can deliver coordinated multimodal responses (that is, joint speech-text response or text-only response}.

The model exploits the autoregressive nature of LLMs by iteratively incorporating both speech tokens \(s_t\) and text tokens \(t_t\) into the Shared LLM Layer at each step. 
These token embeddings are combined through addition to create a unified input representation. The composite embedding \(c_t\) at step \(t\) is formulated as:
\begin{equation}
    c_t = E_{\text{speech}}(s_t) + E_{\text{text}}(t_t)
\end{equation}
\noindent where \(E_{\text{speech}}\) and \(E_{\text{text}}\) represent the embedding functions for speech and text tokens, respectively. To handle the length mismatch between speech and text sequences, we pad the shorter sequence with a special silence token \textless\textbar SIL\textbar\textgreater for each utterance.

The generation follows an autoregressive pattern:
\begin{equation}
    P(y_t | y_{<t}, x) = \prod_{i=1}^t P(y_i | y_{<i}, x)
\end{equation}
\noindent where \(x\) denotes the input and \(y_t = (s_t, t_t)\) represents the combined speech-text output at step \(t\). This formulation \textbf{unifies speech and text generation within one autoregressive process}.

\subsection{Post-Training}
\label{subsec:training_strategy}

Fun-Audio-Chat leverages existing pre-trained models and is trained with a multi-stage post-training pipeline, utilizing millions of hours of diverse speech data that encompasses diverse domains and tasks, including conversational and multilingual speech, audio for understanding tasks, ensuring comprehensive coverage of various scenarios and use cases. The training data combines open-source data following the training setup of DrVoice~\citep{tan2025drvoiceparallelspeechtextvoice} and Audio-Flamingo-3~\citep{DBLP:journals/corr/abs-2507-08128}, along with in-house text, ASR, TTS, audio understanding, speech instruction-following, and voice empathy data.

The multi-stage training pipeline includes: (1) \textbf{Pre-alignment} uses large-scale speech-text paired data for aligning the Speech Encoder, the Adapter, and the Speech Refined Head; (2) \textbf{Core-Cocktail Training}, for supervised full fine-tuning, employs high quality speech data synthesized from billions of text tokens using CosyVoice 3~\citep{DBLP:journals/corr/abs-2505-17589} and selected by thresholding on synthesis Word Error Rate (WER); (3) \textbf{Multi-Task DPO Training} employs diverse real speech data for robustness enhancement, audio understanding and ASR data for comprehension capabilities, instruction-following data (including emotion, style, and prosody control) for speech instruction-following capabilities, and voice empathy data for emotion understanding and empathetic response generation capabilities. This training pipeline is carefully designed to progressively enhance the model's audio comprehension, reasoning, and generation capabilities, while retaining the text capabilities of the backbone LLM.

\paragraph{Pre-Alignment.}
The training process begins with proper initialization of model components. The Speech Encoder is initialized with the weights of Whisper-Large-v3 \citep{radford2022whisper,DBLP:journals/corr/abs-2503-20215}, providing robust voice understanding capabilities. The Shared LLM Layer is initialized using Qwen3-30B-A3B~\citep{DBLP:journals/corr/abs-2505-09388} or alternatively from Vision-language base models Qwen3-VL-8B~\citep{bai2025qwen3vltechnicalreport}, leveraging the strong semantic understanding capabilities of the pre-trained text LLMs. The pre-trained Speech Tokenizer and Detokenizer from CosyVoice 3~\citep{DBLP:journals/corr/abs-2505-17589} are employed and kept frozen throughout the entire training process of Fun-Audio-Chat. To establish effective alignment between audio and text modalities, we perform \textbf{Pre-alignment} training using large-scale speech-text pair data to align the Speech Encoder, the Adapter, and the Speech Refined Head before the main training stages. During this pre-alignment stage, the Shared LLM Layer is kept frozen to preserve its pre-trained capabilities.

\paragraph{Core-Cocktail Training.}
We find that multimodal model training faces a fundamental learning rate trade-off: high learning rates risk degrading the MLLM performance and exacerbating catastrophic forgetting of the base text-LLM's knowledge, while low learning rates cause slow convergence and training stagnation. To address this optimization dilemma and prevent knowledge loss, we utilize the \textbf{Core-Cocktail Training} methodology introduced in our earlier work DrVoice~\citep{tan2025drvoiceparallelspeechtextvoice}, which employs a two-phase training procedure.

\noindent\hspace{1.5em}\textbf{Stage 1: Fine-tuning with High Learning Rate.} In this initial phase, we perform full fine-tuning on all MLLM parameters, the Audio Encoder, and Adapter using an elevated learning rate. For Fun-Audio-Chat, the learning rate is decayed from $1 \times 10^{-4}$ to $1 \times 10^{-5}$ in Stage 1, utilizing a cosine annealing schedule. This stage aims to quickly shift model parameters toward regions of the loss surface that are more conducive to multimodal learning, facilitating rapid task adaptation.

\noindent\hspace{1.5em}\textbf{Intermediate Model Merging.} To mitigate potential MLLM degradation from the intensive Stage 1 training phase, we implement an intermediate model merging operation. Following~\citet{DBLP:conf/acl/XiaoLZX24}, we combine the Stage-1-trained MLLM parameters ($M_1$) with those of the original pretrained LLM ($M_0$) through weighted interpolation, producing a merged model $M_r$:
\begin{equation}
    M_r \leftarrow \alpha M_1 + (1 - \alpha) M_0
\end{equation}
where $\alpha$ controls the interpolation balance. This merging operation reintroduces the foundational knowledge from the base LLM, safeguarding the original text understanding capabilities. Lower $\alpha$ values favor stronger retention of the base LLM's knowledge. In our implementation, $\alpha$ is set to 0.5.

\noindent\hspace{1.5em}\textbf{Stage 2: Refinement with Low Learning Rate.} Stage 2 applies full fine-tuning to the merged model $M_r$ with a reduced learning rate. For Fun-Audio-Chat, the learning rate is decayed from $1 \times 10^{-5}$ to $1 \times 10^{-6}$ in Stage 2, also utilizing a cosine annealing schedule. This enables stable, precise optimization that improves model performance without the instability associated with high learning rates. The Core-Cocktail Training strategy successfully reconciles fast adaptation with knowledge retention, substantially mitigating catastrophic forgetting while promoting effective multimodal learning. Fun-Audio-Chat supports a maximum context length of 2048 tokens (approximately 6 minutes of speech), sufficiently facilitating typical conversational interactions.

\paragraph{Multi-Task DPO Training.}
Following the Core-Cocktail training, we further conduct \textbf{Multi-Task DPO Training}~\citep{DBLP:conf/nips/RafailovSMMEF23} to enhance the model's robustness to real speech data, audio understanding abilities, speech instruction-following and voice empathy capabilities. The Multi-Task DPO Training stage incorporates multiple preference learning objectives: (1) \textbf{robustness preference}: preferring responses that maintain quality under noisy or diverse speech inputs; (2) \textbf{instruction-following preference}: preferring responses that accurately follow voice instructions, including emotion, style, and prosody control; (3) \textbf{audio understanding preference}: preferring responses that demonstrate accurate comprehension of audio content; and (4) \textbf{voice empathy preference}: preferring responses that show appropriate emotional understanding and empathetic responses. The DPO training loss is computed across these multiple preference dimensions, allowing the model to learn a unified preference signal that balances all these capabilities. This multi-task DPO training stage enables the model to better align with human preferences and improve performance on real-world conversational scenarios, distinguishing Fun-Audio-Chat from previous works that primarily rely on supervised fine-tuning.

\paragraph{Full-duplex Interaction Training.}
To enable real-time full-duplex voice interaction, we introduce a parallel speech-text input stream architecture and extend Fun-Audio-Chat to a full-duplex variant, \textbf{Fun-Audio-Chat-Duplex}, which can support natural human-like conversations with seamless two-way communication. Specifically, the parallel speech-text input stream architecture allows the model to accept user speech when the assistant is generating speech, effectively utilizing the time slots that would otherwise be idle. The parallel input stream is designed to handle both user and assistant speech inputs simultaneously, enabling the model to process overlapping speech segments and maintain conversation context. The Full-duplex Interaction Training continues from the checkpoint resulting from the Core-Cocktail Training stage, building upon the multimodal capabilities that the model already acquires. Full-duplex training uses full-duplex conversation data synthesized by augmenting high-quality half-duplex dialogue datasets with simulated full-duplex interaction behaviors, following the data synthesis approach in OmniFlatten~\citep{zhang-etal-2025-omniflatten}. This approach transforms traditional turn-based text dialogues into concurrent dual-stream interactions, in which both user and assistant can speak simultaneously. Full-duplex training allows the model to learn natural turn-taking, interruption handling, and backchanneling behaviors.

\section{Experiments}
\label{sec:experiments}

\subsection{Experimental Setup}
\label{subsec:experimental_setup}

\paragraph{Evaluation Datasets.} 
Following prior works~\citep{yao2024minicpm,kimiteam2025kimiaudiotechnicalreport}, we evaluate the performance of Fun-Audio-Chat comprehensively on the widely used benchmarks:

\begin{itemize}[leftmargin=*,noitemsep]
\item \textbf{Speech-To-Text ($S \rightarrow T$) Evaluation.} We use two types of Spoken Question Answering benchmarks to evaluate the model's ability to understand speech inputs and generate both text and speech responses, including $S \rightarrow T$ Evaluation and $S \rightarrow S$ Evaluation. For $S \rightarrow T$ evaluation, we use VoiceBench~\citep{DBLP:journals/corr/abs-2410-17196} and OpenAudioBench~\footnote{\url{https://huggingface.co/datasets/baichuan-inc/OpenAudioBench}}. \textbf{VoiceBench} encompasses AlpacaEval, CommonEval, SD-QA, MMSU, OpenBookQA, IFEval, and AdvBench, providing comprehensive evaluation across instruction-following, general knowledge, safety alignment, and robustness to real-world variations.
In contrast, \textbf{OpenAudioBench} includes multiple sub-tasks including AlpacaEval~\citep{alpaca_eval}, Llama Q., Reasoning QA, TriviaQA, and Web Q., covering diverse Spoken Question Answering scenarios, with more focuses on general knowledge and reasoning and less emphases on robustness.

\item \textbf{Speech-to-Speech ($S \rightarrow S$) Evaluation.} We use UltraEval-Audio~\footnote{\url{https://github.com/OpenBMB/UltraEval-Audio}}, which includes AlpacaEval, Llama Q., TriviaQA, and Web Q. for end-to-end Speech-to-Speech Question Answering evaluation. 

\item \textbf{Audio Understanding.} We evaluate on \textbf{audio understanding benchmarks} including MMAU~\citep{DBLP:conf/iclr/SakshiTKSSNDGM25}, MMAU-Pro~\citep{DBLP:journals/corr/abs-2508-13992}, and MMSU~\citep{DBLP:journals/corr/abs-2506-04779} for comprehensive audio comprehension capabilities. These benchmarks focus on different aspects of audio understanding: \textbf{MMAU} is a generalist benchmark covering the ``Big Three'' audio domains (Speech, Music, Sound) with a focus on complex reasoning; \textbf{MMAU-Pro} is an advanced-scenario benchmark that stresses models with ``wild'' conditions like long-form audio, spatial audio, and overlapping sounds; \textbf{MMSU} is a speech specialist benchmark grounded in linguistic theory, focusing deeply on the nuances of spoken language (intonation, emotion, prosody) rather than general environmental sounds or music.

\item \textbf{Speech Recognition.} We evaluate ASR performance on the widely used Librispeech~\citep{DBLP:conf/icassp/PanayotovCPK15} for English (EN) ASR and Common Voice~\citep{DBLP:conf/lrec/ArdilaBDKMHMSTW20} for English and Mandarin (ZH) ASR. 

\item \textbf{Speech Function Calling.}  We evaluate on Speech-ACEBench, Speech-BFCL, and Speech-SmartInteract~\footnote{\url{https://huggingface.co/datasets/FunAudioLLM/SpeechFCEval}} for evaluating the model's ability to execute function calls based on speech instructions. These three benchmarks focus on different aspects of speech function calling: \textbf{Speech-ACEBench} is derived from the text-based ACEBench~\citep{DBLP:journals/corr/abs-2501-12851} and contains Mandarin speech recorded by human speakers. It covers both single and parallel function calling scenarios, with particular emphasis on cases where functions take \textit{nested (deep)} object-type arguments. \textbf{Speech-BFCL} is derived from BFCL~\citep{patil2025bfcl} and consists of English data synthesized with TTS. It also targets single and parallel function calling, but focusing on TTS-generated English interactions. \textbf{Speech-SmartInteract} is a purpose-built TTS-synthesized Mandarin speech dataset designed specifically for speech-first interactive use; rather than merely voicing a text-based benchmark, it better reflects the characteristics of real spoken interactions in practical voice assistant settings.

\item \textbf{Speech Instruction-Following and Voice Empathy.} We use the VStyle benchmark~\citep{DBLP:journals/corr/abs-2509-09716} 
to evaluate the model's ability to understand and execute voice instructions for controlling speech generation attributes such as emotion, speaking style, speed, pitch, and volume. We also use an internal test set to assess the model's speech instruction-following and voice empathy capabilities, including understanding emotional context and responding with appropriate empathetic expressions.
\end{itemize}

\paragraph{Evaluation Metrics.} 
Evaluations adhere to the established protocols for each respective benchmark. For $S \rightarrow T$ and $S \rightarrow S$ evaluations on \textbf{Spoken Question Answering benchmarks}, we use different metrics depending on the task type: (1) \textbf{Accuracy} is used for close-ended QA tasks including Llama Q., Reasoning QA, TriviaQA, Web Q., SD-QA, MMSU, OpenBookQA, and IFEval; (2) \textbf{G-Eval}~\citep{DBLP:conf/emnlp/LiuIXWXZ23} is used for open-ended QA tasks including AlpacaEval~\citep{alpaca_eval} and CommonEval, which employs LLM-based evaluation to assess response quality; (3) \textbf{Refusal Rate} is reported for AdvBench to measure safety compliance. 

Additionally, for \textbf{Speech Quality evaluation}, the generated speech is transcribed using Whisper-v3-large model~\citep{radford2022whisper}, then \textbf{ASR-WER} (Word Error Rate of the ASR-transcripts against the model-generated text) is used to assess the alignment between the generated speech and text. \textbf{UTMOS}~\citep{DBLP:conf/interspeech/SaekiXNKTS22} is used to evaluate the overall speech quality, 

For \textbf{Audio Understanding tasks} (MMAU, MMAU-Pro, MMSU), \textbf{Accuracy} is used to measure the model's comprehension capabilities across diverse audio understanding scenarios. For \textbf{Speech Recognition tasks} (Librispeech, Common Voice), \textbf{Word Error Rate (WER)} is reported. 

For \textbf{Speech Function Calling tasks} (Speech-ACEBench, Speech-BFCL, Speech-SmartInteract), \textbf{Accuracy} is used to measure the percentage of correctly executed function calls. 

For \textbf{Speech Instruction-Following and Voice Empathy tasks}, for the VStyle benchmark, we use \textbf{Large Audio Language Model (LALM) evaluation scores} on a 1-5 scale across multiple dimensions: acoustic attributes (age, speed, gender, emotion, pitch, volume), instruction following (emotion, style, variation), role-play (scenario, character), and empathy (anger, sadness, anxiety, joy). For evaluations on our internal test set, similar to the VStyle benchmark, we use LALM as Judge to evaluate the model's performance on Speech Instruction-Following, Semantics-based Empathy, and Paralinguistic-Cue-based Empathy. \textbf{Semantics-based Empathy} refers to the empathy capability that can be judged solely based on text semantics, while \textbf{Paralinguistic-Cue-based Empathy} refers to the empathy capability that requires using Paralinguistic Cues to judge and cannot be judged solely from text semantics. 

For \textbf{Full-Duplex Interaction evaluation}, we use \textbf{S2M-T} (the text output accuracy in multimodal response) and \textbf{S2M-S} (the speech output accuracy in multimodal response) to measure the knowledge understanding performance, and the \textbf{Turn-taking Success Rate} to measure the percentage of interactions where the model correctly handles turn-taking in full-duplex scenarios.

\paragraph{Baselines.} 
We select representative and competitive models as baselines to ensure comprehensive comparisons across different model sizes and model architectures. For \textbf{around-8B dense models}, we compare Fun-Audio-Chat-8B with open-source Large Audio Language Models (LALMs) including GLM-4-Voice (9B)~\citep{DBLP:journals/corr/abs-2412-02612}, MiniCPM-o 2.6 (7B)~\citep{yao2024minicpm}, Baichuan-Omni-1.5 (7B)~\citep{li2025baichuan}, Kimi-Audio (7B)~\citep{kimiteam2025kimiaudiotechnicalreport}, Step-Audio2-Mini (7B)~\citep{DBLP:journals/corr/abs-2507-16632}, and MiMo-Audio (7B)~\citep{coreteam2025mimoaudio}. For \textbf{large-scale models}, we compare Fun-Audio-Chat-30B-A3B with the open-source Longcat-Flash-Omni-Instruct (560B-A27B)~\citep{DBLP:journals/corr/abs-2511-00279} and the closed-source GPT-Audio~\citep{openai2024hello} and Gemini-2.5-Pro~\citep{DBLP:journals/corr/abs-2507-06261}. For \textbf{audio understanding tasks}, we additionally compare with the open-source Audio-Flamingo-3~\citep{DBLP:journals/corr/abs-2507-08128} alongside Kimi-Audio, Step-Audio2-Mini, and MiMo-Audio. For \textbf{speech instruction-following and voice empathy tasks} (VStyle benchmark), we compare with the open-source Baichuan-Audio~\citep{DBLP:journals/corr/abs-2502-17239} and Kimi-Audio, and add the closed-source GPT-4o~\citep{openai2024hello} and Doubao~\footnote{\url{https://www.doubao.com/}} into baselines for comprehensive comparison with both open-source and commercial models. For \textbf{full-duplex interaction evaluation}, we compare Fun-Audio-Chat-Duplex with the open-source Moshi~\citep{kyutai2024moshi} and FreezeOmni~\citep{DBLP:conf/icml/WangLFZS000M25}. 

In summary, the selected baselines cover diverse modeling paradigms (Text-Driven vs. Joint Speech-Text, interleaved vs. parallel architectures) and model scales, enabling systematic comparisons across mainstream speech-text modeling strategies and providing comprehensive evaluation of Fun-Audio-Chat's capabilities across different task categories.

\subsection{Spoken Question Answering}
\label{subsec:spoken_qa}

\paragraph{Accuracy.} Fun-Audio-Chat demonstrates strong performance on spoken question answering tasks. Table~\ref{tab:speech_qa_performance_large} compares Fun-Audio-Chat-30B-A3B with large-scale baselines, including GPT-Audio, Gemini-2.5-Pro, and Longcat-Flash-Omni-Instruct. Table~\ref{tab:speech_qa_performance} compares Fun-Audio-Chat-8B with Kimi-Audio, Step-Audio2-Mini, and other similarly-scaled open-source models. As shown in Table~\ref{tab:speech_qa_performance_large} and Table~\ref{tab:speech_qa_performance}, Fun-Audio-Chat achieves competitive performance among similarly-scaled models (8B and 30B-A3B parameters). Specifically, \textbf{Fun-Audio-Chat-8B achieves the best overall performance on OpenAudioBench (76.61\%) and VoiceBench (83.21\%) among $\sim$8B-scale models}, while \textbf{Fun-Audio-Chat-30B-A3B achieves competitive results compared to large-scale baselines, including top-tier closed-source models}.

\begin{table}[t]
    \centering
    \caption{Performance comparison on \textbf{Spoken Question Answering} benchmarks for \textbf{large-scale models}. The best result in each row is in \textbf{bold}. \textbf{Frame Rate-In} denotes the input speech frame rate (Hz), and \textbf{Frame Rate-Out} denotes the output (speech + text) frame rate (Hz) for the LLM backbone. }
    \label{tab:speech_qa_performance_large}
    \centering
    \small
    \begin{tabular}{lcccc}
    \toprule
    \textbf{} & \textbf{GPT-Audio} & \textbf{Gemini-2.5-Pro} & \makecell{\textbf{Longcat-Flash} \\ \textbf{-Omni-Instruct}} & \makecell{\textbf{Fun-Audio-Chat} \\ \textbf{-30B-A3B}} \\
    \midrule
    LLM Size & -- & -- & 560B-A27B & 30B-A3B \\
    Frame Rate-In & -- & -- & 12.5 & \textbf{5} \\
    Frame Rate-Out & -- & -- & 16.67 & \textbf{5} \\
    \midrule
    \multicolumn{5}{c}{\textit{OpenAudioBench (S2T)}} \\
    \midrule
    AlpacaEval          & 83.37 & 76.58 & 75.43 & \textbf{88.89} \\
    Llama Q.             & \textbf{90.67} & 83.00 & 83.33 & 85.00 \\
    Reasoning QA   & 74.75 & \textbf{80.30} & 79.71 & 75.25 \\
    TriviaQA            & \textbf{92.20} & 90.20 & 86.20 & 76.00 \\
    Web Q.              & \textbf{83.70} & 80.90 & 76.00 & 77.80 \\
    Overall               & \textbf{84.94} & 82.20 & 80.13 & 80.59 \\
    \midrule
    \multicolumn{5}{c}{\textit{VoiceBench (S2T)}} \\
    \midrule
    AlpacaEval          & 4.84 & 4.70 & \textbf{4.94} & 4.82  \\
    CommonEval          & 4.47 & 4.11 & 4.32 & \textbf{4.49}  \\
    SD-QA               & \textbf{89.72} & 83.54 & 82.46 & 72.87 \\
    MMSU                & 83.25 & \textbf{88.32} & 81.95 & 75.31 \\
    OpenBookQA          & 92.53 & \textbf{95.16} & 93.41 & 88.57 \\
    IFEval              & \textbf{79.12} & 77.83 & 77.99 & 77.25 \\
    AdvBench            & 99.62 & 97.69 & \textbf{100} & 99.23 \\
    Overall             & \textbf{90.06} & 88.39 & 88.72 & 85.63 \\
    \midrule
    \multicolumn{5}{c}{\textit{UltraEval-Audio (S2S)}} \\
    \midrule
    AlpacaEval          & \textbf{73.38} & -- & -- & 64.49     \\
    Llama Q.            & \textbf{89.00} & -- & -- & 78.67 \\
    TriviaQA            & \textbf{72.85} & -- & -- & 54.20     \\
    Web Q.              & \textbf{55.41} & -- & -- & 51.18     \\
    Overall             & \textbf{72.66} & -- & -- & 62.14     \\
    \bottomrule
    \end{tabular}
    \end{table}
    
    \begin{table}[t]
    \centering
    \caption{Performance comparison on \textbf{Spoken Question Answering} benchmarks for \textbf{$\sim$8B-scale dense models}. The best result in each row is in \textbf{bold}. \textbf{Frame Rate-In} denotes the input speech frame rate (Hz), and \textbf{Frame Rate-Out} denotes the output (speech + text) frame rate (Hz) for the LLM backbone. $\tau$ denotes the average number of text tokens per second of speech.}
    \label{tab:speech_qa_performance}
    \centering
    \small
    \begin{tabular}{lcccccccc}
    \toprule
    \textbf{} & \makecell{\textbf{GLM4} \\ \textbf{-Voice}} & \makecell{\textbf{MiniCPM} \\ \textbf{-o 2.6}} & \makecell{\textbf{Baichuan} \\ \textbf{-Omni-1.5}} & \makecell{\textbf{Kimi} \\ \textbf{-Audio}} & \makecell{\textbf{Step-Audio2} \\ \textbf{-Mini}} & \makecell{\textbf{MiMo} \\ \textbf{-Audio}} & \makecell{\textbf{Fun-Audio-Chat} \\ \textbf{-8B}} \\
    \midrule
    LLM Size &  9B & 7B & 7B & 7B & 7B & 7B & 8B \\
    Frame Rate-In & 12.5 & 25 & 12.5 & 12.5 & 12.5 & 6.25 & \textbf{5} \\
    Frame Rate-Out & 12.5+${\tau}$ & ${\tau}$ & 12.5+${\tau}$ & 12.5 & 25+${\tau}$ & 6.25+${\tau}$ & \textbf{5} \\
    \midrule
    \multicolumn{8}{c}{\textit{OpenAudioBench (S2T)}} \\
    \midrule
    AlpacaEval          & 57.89 & 64.10 & 77.90 & 75.73 & 59.60 & 85.43 & \textbf{88.94} \\
    Llama Q.             & 76.00 & 78.00 & 78.50 & 79.33 & 75.00 & 79.67 & \textbf{83.33} \\
    Reasoning QA   & 47.43 & 38.60 & 50.00 & 58.02 & 46.04 & 53.96 & \textbf{69.80} \\
    TriviaQA            & 51.80 & 63.00 & 57.20 & 62.10 & 57.70 & 52.80 & \textbf{68.10} \\
    Web Q.              & 55.40 & 69.20 & 59.10 & 70.20 & 65.10 & 55.40 & \textbf{72.90} \\
    Overall               & 57.70 & 62.58 & 64.54 & 69.08 & 60.69 & 65.45 & \textbf{76.61} \\
    \midrule
    \multicolumn{8}{c}{\textit{VoiceBench (S2T)}} \\
    \midrule
    AlpacaEval          & 3.97 & 4.42 & 4.50 & 4.46 & 4.17 & 4.60 & \textbf{4.80}  \\
    CommonEval          & 3.42 & 4.15 & 4.05 & 3.97 & 3.00 & 3.77 & \textbf{4.42}  \\
    SD-QA               & 36.98 & 50.72 & 43.40 & 63.12 & 56.06 & 54.79 & \textbf{66.27} \\
    MMSU                & 39.75 & 54.78 & 57.25 & 62.17 & 52.18 & 59.66 & \textbf{71.08} \\
    OpenBookQA          & 53.41 & 78.02 & 74.51 & \textbf{83.52} & 64.18 & 73.41 & \textbf{83.52} \\
    IFEval              & 52.80 & 49.25 & 54.54 & 61.10 & 38.01 & 66.45 & \textbf{78.52} \\
    AdvBench            & 88.08 & 97.69 & 97.31 & \textbf{100.00} & 93.08 & 96.73 & 98.65 \\
    Overall             & 59.83 & 71.69 & 71.14 & 76.93 & 63.84 & 74.06 & \textbf{83.21} \\
    \midrule
    \multicolumn{8}{c}{\textit{UltraEval-Audio (S2S)}} \\
    \midrule
    AlpacaEval          & 51.00 & 51.00 & 58.69 & 44.20 & 51.72 & 61.46 & \textbf{61.87}     \\
    Llama Q.              & 50.00 & 61.00 & 67.33 & 57.33 & 67.67 & 77.33 & \textbf{78.33} \\
    TriviaQA            & 36.40 & 40.20 & 30.57 & 35.71 & 33.50 & 40.43 & \textbf{49.51}     \\
    Web Q.                & 32.00 & 40.00 & 38.09 & 33.90 & 34.65 & 42.86 & \textbf{48.52}     \\
    Overall               & 42.35 & 48.05 & 48.67 & 42.79 & 46.89 & 55.52 & \textbf{59.56}     \\
    \bottomrule
    \end{tabular}
    \end{table}

\paragraph{Speech Quality.}
We evaluate the speech quality of Fun-Audio-Chat-8B on UltraEval-Audio using UTMOS for the overall speech quality and ASR-WER for alignment between the generated speech and text. On the Llama Q. test set, Fun-Audio-Chat-8B achieves a UTMOS score of 4.37, indicating excellent overall speech quality, and an ASR-WER of 4.32\%, demonstrating strong alignment between the generated speech and the corresponding text outputs. These results demonstrate that the dual-resolution architecture maintains high-quality speech generation despite operating at the efficient 5Hz frame rate, validating the effectiveness of the Dual-Resolution Speech Representations (DRSR) architecture in balancing efficiency and speech quality. 

\subsection{Audio Understanding}
\label{subsec:audio_understanding}
\vspace{-2mm}

Table~\ref{tab:audio_understanding} demonstrates that \textbf{Fun-Audio-Chat achieves the best performance on comprehensive audio understanding benchmarks including MMAU, MMAU-Pro, and MMSU, over strong open-source baselines}, including Kimi-Audio~\citep{kimiteam2025kimiaudiotechnicalreport}, Audio-Flamingo-3~\citep{DBLP:journals/corr/abs-2507-08128}, MiMo-Audio~\citep{coreteam2025mimoaudio}, and Step-Audio2-Mini~\citep{DBLP:journals/corr/abs-2507-16632}. On MMAU~\footnote{MMAU v05.15.25 test-mini.}, Fun-Audio-Chat-30B-A3B achieves the best performance (77.9\%) among all evaluated models, followed by Fun-Audio-Chat-8B (76.6\%). On MMAU-Pro~\footnote{Only one audio test case included.}, Fun-Audio-Chat-30B-A3B achieves the best result (59.9\%), with Fun-Audio-Chat-8B achieving the second-best performance (58.0\%). On MMSU, Fun-Audio-Chat-30B-A3B achieves 70.1\%, the highest result among all models, followed by Fun-Audio-Chat-8B (67.8\%). For speech recognition tasks, Fun-Audio-Chat achieves competitive WERs across multiple datasets in both English (EN) and Mandarin (ZH), demonstrating robust audio comprehension capabilities across diverse domains and languages.

\begin{table}[t]
    \centering
    \caption{Performance comparison on \textbf{Audio understanding} (top section) on MMAU, MMAU-Pro, and MMSU, and \textbf{Speech Recognition} (bottom) on Librispeech and Common Voice. The best result in each row is in \textbf{bold}.}
    \label{tab:audio_understanding}
    \centering
    \small
    \begin{tabular}{lcccccc}
    \toprule
    \textbf{} & \makecell{\textbf{Kimi} \\ \textbf{-Audio}} & \makecell{\textbf{Audio} \\ \textbf{-Flamingo-3}} & \makecell{\textbf{Step-Audio2} \\ \textbf{-Mini}} & \makecell{\textbf{MiMo} \\ \textbf{-Audio}} & \makecell{\textbf{Fun-Audio-Chat} \\ \textbf{-30B-A3B}} & \makecell{\textbf{Fun-Audio-Chat} \\ \textbf{-8B}} \\
    \midrule
    \multicolumn{7}{c}{\textit{Audio Understanding}} \\
    \midrule
    MMAU & 69.6 & 73.3 & 73.2 & 74.9 & \textbf{77.9} & 76.6 \\
    MMAU-Pro & 46.6 & 51.7 & 53.2 & 53.4 & \textbf{59.9} & 58.0 \\
    MMSU & 59.3 & 61.4 & 56.8 & 61.7 & \textbf{70.1} & 67.8 \\
    \midrule
    \multicolumn{7}{c}{\textit{Speech Recognition}} \\
    \midrule
    Librispeech clean         & \textbf{1.28} & 1.57 & 1.33 & 3.56 & 1.64 & 1.71 \\
    Librispeech other         & \textbf{2.42} & 3.13 & 2.86 & 16.22 & 3.73 & 4.13 \\
    Common Voice-EN      & 10.31 & 7.4 & \textbf{6.76} & 62.05 & 7.79 & 8.88 \\
    Common Voice-ZH      & 7.21 & -- & \textbf{5.38} & 44.11 & 5.88 & 6.16 \\
    \bottomrule
    \end{tabular}
    \end{table}

\subsection{Speech Function Calling}
\label{subsec:speech_function_calling}
\vspace{-2mm}

Table~\ref{tab:function_call} presents the performance of Fun-Audio-Chat on speech function calling benchmarks.  \textbf{Fun-Audio-Chat-30B-A3B achieves the highest overall score (79.63\%) among all evaluated models}, with particularly strong performance on Speech-ACEBench (Single: 76.40\%) and Speech-SmartInteract (84.13\%). The model demonstrates strong capabilities in understanding speech-based function calling instructions and executing them accurately, which is crucial for building practical voice-controlled applications. \textbf{The performance on parallel function calling scenarios (54.50\% on ACEBench-Parallel and 87.63\% on BFCL-Parallel by Fun-Audio-Chat-8B) further highlights Fun-Audio-Chat's ability to handle complex, multi-step instructions in voice interactions, with Fun-Audio-Chat-8B outperforming the top tier closed-source GPT-Audio and Gemini-2.5-Pro on BFCL-Parallel}.

\begin{table}[t]
\centering
\caption{Performance comparison on \textbf{Speech Function Calling}. The best result in each row is in \textbf{bold}.}
\label{tab:function_call}
\centering
\small
\begin{tabular}{lccccc}
\toprule
\textbf{} & \textbf{GPT-Audio} &\makecell{\textbf{Gemini} \\ \textbf{-2.5-Pro}}  & \makecell{\textbf{Step-Audio2} \\ \textbf{-Mini}} & \makecell{\textbf{Fun-Audio-Chat} \\ \textbf{-30B-A3B}} & \makecell{\textbf{Fun-Audio-Chat} \\ \textbf{-8B}} \\
\midrule
Speech-ACEBench (Single) & 68.30 & 68.30 & 38.90 & \textbf{76.40} & 66.30 \\
Speech-ACEBench (Parallel) & \textbf{60.20} & 53.40 & 4.50 & 59.10 & 54.50 \\
Speech-BFCL (Single) & 88.58 & 88.41 & 77.51 & 92.21 & \textbf{92.73} \\
Speech-BFCL (Parallel) & 83.60 & 80.91 & 49.73 & 86.29 & \textbf{87.63} \\
Speech-SmartInteract (Single) & 66.77 & 79.19 & 41.92 & \textbf{84.13} & 79.79 \\
\midrule
Overall & 73.49 & 74.04 & 42.51 & \textbf{79.63} & 76.19 \\
\bottomrule
\end{tabular}
\end{table}

\subsection{Speech Instruction-Following and Voice Empathy}
\label{subsec:speech_instruction_following}
\vspace{-2mm}

Table~\ref{tab:vstyle} and Table~\ref{tab:empathy} demonstrate that \textbf{Fun-Audio-Chat achieves strong performance on Speech Instruction-Following and Voice Empathy tasks}.  As shown in Table~\ref{tab:vstyle}, Fun-Audio-Chat-30B-A3B and Fun-Audio-Chat-8B demonstrate competitive performance on Speech Instruction-Following across multiple dimensions, including acoustic attributes, instruction following, role-play, and empathy capabilities, in both English and Chinese, substantially outperforming open-source models including Baichuan-Audio and Kimi-Audio while remaining competitive with commercial models. (1) In terms of the \textbf{Overall} performance, Fun-Audio-Chat-8B achieves scores 3.35 and 3.46 for English and Mandarin respectively, substantially outperforming the open-source Baichuan-Audio (2.50/2.25) and Kimi-Audio (2.54/3.11) in both languages, while remaining competitive with commercial models. (2) Specifically, for \textbf{acoustic attributes}, Fun-Audio-Chat-8B shows strong performance in emotion control (4.13/4.00 for en/zh) and volume control (3.95/3.70), demonstrating effective acoustic attribute manipulation capabilities. Notably, Fun-Audio-Chat-8B achieves the best performance on Age control in English (4.04), and achieves a score of 4.20 on speed control in Mandarin, ranking second only to Doubao (4.35). (3) In \textbf{instruction-following} tasks, Fun-Audio-Chat-8B achieves moderate performance with scores of 4.09 and 3.14 for style control in English and Mandarin, indicating room for improvements in complex instruction understanding. (4) For \textbf{role-play} capabilities, Fun-Audio-Chat-8B performs better in Mandarin (3.42/3.30 for scenario/character) compared to English (2.50/3.06), suggesting stronger contextual understanding in Mandarin scenarios.

\begin{table}[t]
    \centering
    \caption{Performance comparison on \textbf{Speech Instruction-Following} on the VStyle benchmark~\citep{DBLP:journals/corr/abs-2509-09716}. The best result in each row for each language is in \textbf{bold}.}
    \label{tab:vstyle}
    \centering
    \small
    \begin{tabular}{lcccccccc}
    \toprule
    \textbf{} & \textbf{Lang} & \makecell{\textbf{GPT} \\ \textbf{-Audio}} & \makecell{\textbf{GPT} \\ \textbf{-4o}} & \textbf{Doubao} & \makecell{\textbf{Baichuan} \\ \textbf{-Audio}} & \makecell{\textbf{Kimi} \\ \textbf{-Audio}} & \makecell{\textbf{Fun-Audio-Chat} \\ \textbf{-30B-A3B}} & \makecell{\textbf{Fun-Audio-Chat} \\ \textbf{-8B}} \\
    \midrule
    \multirow{2}{*}{Overall} & en & 3.78 & \textbf{4.05} & 3.63 & 2.50 & 2.54 & 3.31 & 3.35 \\
    & zh & 3.75 & 3.84 & \textbf{4.10} & 2.25 & 3.11 & 3.68 & 3.46 \\
    \midrule
    \multicolumn{9}{c}{\textit{Acoustic Attributes}} \\
    \midrule
    \multirow{2}{*}{Age} & en & 3.67 & 3.67 & 3.75 & 2.71 & 2.79 & 3.79 & \textbf{4.04} \\
    & zh & 3.67 & 3.42 & \textbf{3.88} & 2.67 & 3.33 & 3.42 & 3.50 \\
    \multirow{2}{*}{Speed} & en & \textbf{4.05} & 3.45 & 3.55 & 2.20 & 2.45 & 3.55 & 3.45 \\
    & zh & 3.65 & 3.10 & 4.35 & 2.45 & 3.45 & \textbf{4.47} & 4.20 \\
    \multirow{2}{*}{Gend.} & en & 3.75 & 2.79 & 3.46 & \textbf{3.83} & 2.54 & 2.96 & 3.33 \\
    & zh & \textbf{4.08} & 3.50 & 3.25 & 3.08 & 2.25 & 3.26 & 3.08 \\
    \multirow{2}{*}{Emot.} & en & \textbf{4.50} & 4.00 & 3.38 & 2.58 & 3.04 & 4.25 & 4.13 \\
    & zh & 4.42 & 3.83 & \textbf{4.65} & 2.29 & 3.75 & 4.08 & 4.00 \\
    \multirow{2}{*}{Pitch} & en & 3.30 & \textbf{3.60} & 3.25 & 2.05 & 1.55 & 2.95 & 3.20 \\
    & zh & 3.05 & 3.35 & \textbf{4.35} & 2.00 & 2.95 & 3.00 & 2.75 \\
    \multirow{2}{*}{Vol.} & en & \textbf{4.20} & 4.10 & 4.05 & 2.05 & 3.00 & 3.90 & 3.95 \\
    & zh & 4.25 & 3.90 & \textbf{4.70} & 2.80 & 3.25 & 4.35 & 3.70 \\
    \multirow{2}{*}{Comp.} & en & \textbf{3.73} & 3.27 & 3.13 & 2.55 & 2.33 & 3.36 & 3.17 \\
    & zh & 3.47 & 3.22 & \textbf{3.77} & 2.58 & 3.17 & 3.60 & 3.37 \\
    \midrule
    \multicolumn{9}{c}{\textit{Instruction}} \\
    \midrule
    \multirow{2}{*}{Emot.} & en & \textbf{4.13} & 3.93 & 3.52 & 2.23 & 2.19 & 3.88 & 3.70 \\
    & zh & 3.73 & 3.37 & \textbf{3.90} & 1.71 & 2.66 & 3.66 & 3.42 \\
    \multirow{2}{*}{Style} & en & \textbf{4.51} & 4.23 & 3.67 & 2.21 & 2.41 & 3.79 & 4.09 \\
    & zh & \textbf{4.07} & 3.51 & 3.96 & 1.72 & 2.74 & 4.01 & 3.14 \\
    \multirow{2}{*}{Vari.} & en & 4.03 & \textbf{4.07} & 2.90 & 1.88 & 2.33 & 3.47 & 3.06 \\
    & zh & \textbf{3.48} & 3.11 & 2.88 & 1.69 & 2.43 & 2.96 & 2.94 \\
    \midrule
    \multicolumn{9}{c}{\textit{Role-Play}} \\
    \midrule
    \multirow{2}{*}{Scen.} & en & 2.65 & \textbf{3.89} & 3.27 & 2.08 & 1.73 & 2.65 & 2.50 \\
    & zh & 3.69 & 3.89 & \textbf{4.45} & 2.29 & 3.01 & 4.02 & 3.42 \\
    \multirow{2}{*}{Char.} & en & 3.37 & \textbf{3.83} & 2.56 & 2.33 & 1.72 & 2.48 & 3.06 \\
    & zh & 3.65 & 3.90 & 3.79 & 1.95 & 2.23 & \textbf{3.95} & 3.30 \\
    \midrule
    \multicolumn{9}{c}{\textit{Empathy}} \\
    \midrule
    \multirow{2}{*}{Anger} & en & 4.25 & \textbf{4.95} & 4.89 & 2.41 & 3.59 & 2.84 & 3.64 \\
    & zh & 3.80 & \textbf{4.75} & 4.59 & 2.11 & 3.86 & 3.64 & 3.73 \\
    \multirow{2}{*}{Sad.} & en & 3.80 & 4.90 & \textbf{5.00} & 3.43 & 3.97 & 4.00 & 4.10 \\
    & zh & 3.62 & \textbf{4.83} & 4.72 & 2.55 & 3.86 & 3.83 & 3.93 \\
    \multirow{2}{*}{Anx.} & en & 4.23 & \textbf{5.00} & 4.81 & 2.74 & 3.65 & 3.61 & 2.90 \\
    & zh & 4.33 & 4.67 & \textbf{4.80} & 2.20 & 3.80 & 2.90 & 4.03 \\
    \multirow{2}{*}{Joy} & en & 3.97 & 4.54 & \textbf{4.94} & 3.91 & 3.46 & 3.54 & 3.69 \\
    & zh & 3.91 & 4.80 & \textbf{4.83} & 3.51 & 4.57 & 3.71 & 3.77 \\
    \bottomrule
    \end{tabular}
    \end{table}

We further evaluate Speech Instruction-Following and Voice Empathy capabilities on our internal test set, as shown in Table~\ref{tab:empathy}. Notably, Fun-Audio-Chat achieves superior performance over GPT-Audio in terms of both Semantics-based Empathy and Paralinguistic-Cue-based Empathy, demonstrating \textbf{the model's strong ability to understand emotional context and respond with appropriate empathetic expressions}.

\begin{table}[t]
    \centering
    \caption{Performance comparison on \textbf{Speech Instruction-Following and Voice Empathy} on our internal test set. The best result in each row is in \textbf{bold}.}
    \label{tab:empathy}
    \centering
    \small
    \begin{tabular}{lccc}
    \toprule
    & \textbf{GPT-Audio} & \makecell{\textbf{Fun-Audio-Chat} \\ \textbf{-30B-A3B}} & \makecell{\textbf{Fun-Audio-Chat} \\ \textbf{-8B}} \\
    \midrule
    Speech Instruction-Following & \textbf{4.53} & 4.31 & 3.98 \\
    Semantics-based Empathy & 4.73 & \textbf{4.80} & \textbf{4.80} \\
    Paralinguistic-Cue-based Empathy & 3.20 & 3.55 & \textbf{3.85} \\
    \bottomrule
    \end{tabular}
    \end{table}

\subsection{Full-Duplex Interaction}
We evaluate the full-duplex variant Fun-Audio-Chat-Duplex on two key aspects: knowledge understanding in full-duplex scenarios and objective full-duplex interaction metrics.

\noindent \textbf{Full-Duplex Knowledge Understanding.}
Table~\ref{tab:duplex_knowledge_turntaking} shows the full-duplex knowledge understanding performance of Fun-Audio-Chat-Duplex. The results demonstrate that \textbf{Fun-Audio-Chat-Duplex maintains strong knowledge understanding capabilities in full-duplex conversation scenarios}. Fun-Audio-Chat-Duplex-30B-A3B achieves the highest average performance on both S2M-T (54.89\%) and S2M-S (49.28\%) metrics, significantly outperforming Moshi (33.17\%/29.86\%) and FreezeOmni~\citep{DBLP:conf/icml/WangLFZS000M25} (47.58\%/34.49\%). On individual benchmarks, Fun-Audio-Chat-Duplex-30B-A3B achieves the highest results on Llama Q. (81.00\%/71.33\%), AlpacaEval (68.23\%/59.65\%), and TriviaQA (41.70\%/40.04\%) for both text and speech outputs. This indicates that the full-duplex architecture successfully preserves the model's knowledge comprehension abilities while enabling simultaneous two-way communication, allowing the system to maintain context and understanding even when processing overlapping speech inputs and outputs.

\noindent \textbf{Full-Duplex Interaction.}
Table~\ref{tab:duplex_knowledge_turntaking} also presents the turn-taking success rates for full-duplex voice interactions. Fun-Audio-Chat-Duplex-30B-A3B achieves perfect turn-taking success rate (100.00\%), outperforming both Moshi (99.77\%) and FreezeOmni~\citep{DBLP:conf/icml/WangLFZS000M25} (93.87\%). Fun-Audio-Chat-Duplex-8B achieves 99.94\%, also demonstrating excellent turn-taking capabilities. These results indicate that Fun-Audio-Chat-Duplex successfully enables natural and efficient full-duplex voice interactions, with the model's ability to handle simultaneous speech and maintain appropriate conversation flow, closely mirroring the dynamics of human-human conversations.

\begin{table}[t]
    \centering
    \caption{Performance comparison on Knowledge Understanding (in terms of \textbf{S2M-T} (text output in multimodal response) and \textbf{S2M-S} (speech output in multimodal response)) and Full-duplex Interaction (in terms of Turn-taking Success Rate) on the full-duplex variant of the UltraEvalAudio benchmark. The best result for each metric on each dataset is in \textbf{bold}.}
    \label{tab:duplex_knowledge_turntaking}
    \small
    \begin{tabular}{lcccccccccc}
    \toprule
    \multirow{2}{*}{\textbf{}} & \multicolumn{2}{c}{\textbf{Moshi}} & \multicolumn{2}{c}{\textbf{FreezeOmni}} & \multicolumn{2}{c}{\makecell{\textbf{Fun-Audio-Chat} \\ \textbf{-Duplex-30B-A3B}}} & \multicolumn{2}{c}{\makecell{\textbf{Fun-Audio-Chat} \\ \textbf{-Duplex-8B}}} \\
    \cmidrule(lr){2-3} \cmidrule(lr){4-5} \cmidrule(lr){6-7} \cmidrule(lr){8-9}
    & \textbf{S2M-T} & \textbf{S2M-S} & \textbf{S2M-T} & \textbf{S2M-S} & \textbf{S2M-T} & \textbf{S2M-S} & \textbf{S2M-T} & \textbf{S2M-S} \\
    \midrule
    Llama Q. & 65.67 & 57.00 & 74.00 & 58.00 & \textbf{81.00} & \textbf{71.33} & 72.33 & 64.33 \\
    AlpacaEval & 25.51 & 25.08 & 47.39 & 32.78 & \textbf{68.23} & \textbf{59.65} & 68.03 & 57.32 \\
    TriviaQA & 18.46 & 16.31 & 30.08 & 20.31 & \textbf{41.70} & \textbf{40.04} & 29.59 & 27.73 \\
    Web Q. & 23.03 & 21.06 & \textbf{38.83} & \textbf{26.87} & 28.64 & 26.08 & 26.18 & 24.36 \\
    \midrule
    Avg. & 33.17 & 29.86 & 47.58 & 34.49 & \textbf{54.89} & \textbf{49.28} & 49.03 & 43.44 \\
    \midrule
    Turn-taking Success Rate & \multicolumn{2}{c}{99.77} & \multicolumn{2}{c}{93.87} & \multicolumn{2}{c}{\textbf{100.00}} & \multicolumn{2}{c}{99.94} \\
    \bottomrule
    \end{tabular}
    \end{table}

\subsection{Computational Efficiency}

A key advantage of Fun-Audio-Chat is its computational efficiency, highlighted in Table~\ref{tab:speech_qa_performance_large} and Table~\ref{tab:speech_qa_performance}. As shown in the \textbf{Frame Rate-In/Frame Rate-Out} rows, Fun-Audio-Chat operates at a frame rate of \textbf{5/5 Hz}, indicating that the LLM backbone processes only 5 audio tokens per second for both input and output. This represents a 1.25$\times$ to 5$\times$ reduction in input frame rate compared to other models, which operate at frame rates ranging from 6.25Hz (MiMo-Audio) to 25Hz (MiniCPM-o 2.6), with most models using 12.5Hz. For output frame rates, Fun-Audio-Chat's 5Hz is significantly lower than other models, which operate at rates of 12.5Hz, 16.67Hz, 25Hz, or higher when including text token generation, e.g., 12.5+$\tau$ for GLM-4-Voice and Baichuan-Omni-1.5, 25+$\tau$ for Step-Audio2-Mini, where $\tau$ denotes the average number of text tokens per second of speech. \textbf{The dual-resolution design significantly reduces computational requirements and potential latency, with empirical measurements showing approximately 50\% reduction in GPU hours during training compared to models operating at higher frame rates}. Importantly, this efficiency is achieved without compromising speech quality, as demonstrated by the high-quality speech generation results.

\section{Conclusion}
\label{sec:conclusion}

This report introduces Fun-Audio-Chat, a large-scale Large Audio Language Model (LALM) designed to overcome the limitations of existing joint speech-text models for seamless voice interaction. Fun-Audio-Chat extends our previous work DrVoice~\citep{tan2025drvoiceparallelspeechtextvoice} by adopting one key innovation, Dual-Resolution Speech Representations (DRSR) architecture, at significantly larger scales. The DRSR architecture enables the Shared LLM backbone to process audio at an efficient 5Hz frame rate (Frame Rate-In/Frame Rate-Out: 5/5 Hz) while the Speech Refined Head generates high-quality speech tokens at 25Hz resolution. This dual-resolution design effectively balances computational efficiency (reducing GPU hours by nearly 50\%) and speech generation quality.

To address the catastrophic forgetting challenge in multimodal learning, we adopt the Core-Cocktail Training strategy introduced in DrVoice~\citep{tan2025drvoiceparallelspeechtextvoice}, a two-stage approach with intermediate parameter merging. Subsequently, we enhance the model through Multi-Task DPO Training to strengthen the robustness to real speech data, capabilities of speech instruction-following, audio understanding, and voice empathy. The multi-stage post-training paradigm enables Fun-Audio-Chat to retain the original text-LLM's capabilities while gaining powerful multimodal skills.

Trained on millions of hours of diverse speech data and scaled to larger model sizes (dense 8B and MoE 30B-A3B parameters), Fun-Audio-Chat achieves strong performance on Spoken Question Answering (Speech-to-Text and Speech-to-Speech generation) tasks, ranking Top among models of the same sizes. It also achieves competitive results on audio understanding, speech function calling, speech instruction-following, and voice empathy tasks, as demonstrated across comprehensive benchmarks including OpenAudioBench, VoiceBench, UltraEvalAudio, MMAU, MMAU-Pro, MMSU, Speech-ACEBench, Speech-BFCL, Speech-SmartInteract, and VStyle. Furthermore, we develop Fun-Audio-Chat-Duplex, a full-duplex variant that achieves strong performance on Spoken Question Answering benchmarks and full-duplex interactions.

We open-source Fun-Audio-Chat-8B model, including the model checkpoint and its training and inference code, and provide an interactive demo, encouraging researchers and practitioners to experience and build upon our work. We believe that Fun-Audio-Chat represents a significant advancement in the field of voice interaction systems, demonstrating that carefully designed large-scale post-training and architectural innovations can significantly enhance the audio comprehension, reasoning, and speech generation capabilities of LALMs while achieving high computational efficiency.

\section{Limitations}

While Fun-Audio-Chat demonstrates strong performance across multiple benchmarks, several limitations remain to be addressed in future work. First, for complex question answering in multi-turn conversations, the model occasionally exhibits memory loss of context, where information from earlier turns may not be consistently retained. This limitation is particularly noticeable in scenarios requiring long-context comprehension and complex reasoning across multiple turns.

Second, speech instruction-following capabilities show some instability in expressiveness. While the model generally performs strongly on voice instruction tasks, there are cases where the generated speech may not fully capture the intended emotional nuances, speaking styles, or prosodic variations specified in the instructions. This variability in expressiveness can affect the naturalness and appropriateness of voice responses in certain contexts.

Third, the voice empathy capabilities demonstrate some instability in performance. Although Fun-Audio-Chat achieves competitive results on empathy evaluation benchmarks (including both Semantics-based Empathy and Paralinguistic-Cue-based Empathy), the model's ability to consistently recognize and respond with appropriate emotional empathy can vary across different scenarios and emotional contexts. This inconsistency may impact the reliability of empathetic response generation in real-world applications where emotional understanding is critical.

These limitations highlight important directions for future research, including improving long-term context management in multi-turn conversations, enhancing the stability and expressiveness of speech instruction-following, and developing more robust and consistent voice empathy capabilities across diverse emotional scenarios.

\section{Contributions and Acknowledgments}
All contributors of Fun-Audio-Chat are listed in alphabetical order by their last names.
\noindent\paragraph{Core contributors:} {Qian Chen}, {Luyao Cheng}, {Chong Deng}, {Xiangang Li}, {Jiaqing Liu}, {Chao-Hong Tan}, {Wen Wang}, {Junhao Xu}, {Jieping Ye}, {Qinglin Zhang}, {Qiquan Zhang}, {Jingren Zhou}
\noindent\paragraph{Contributors:} {Zhifu Gao}, {Weiqin Li}, {Mengge Liu}, {Xiang Lv}, {Yukun Ma}, {Gang Qiao}, {Hui Wang}, {Chong Zhang}, {Han Zhao}, {Tianyu Zhao}

\clearpage

\bibliography{biblio}

\begin{thebibliography}{43}
\providecommand{\natexlab}[1]{#1}
\providecommand{\url}[1]{\texttt{#1}}
\expandafter\ifx\csname urlstyle\endcsname\relax
  \providecommand{\doi}[1]{doi: #1}\else
  \providecommand{\doi}{doi: \begingroup \urlstyle{rm}\Url}\fi

\bibitem[Ardila et~al.(2020)Ardila, Branson, Davis, Kohler, Meyer, Henretty, Morais, Saunders, Tyers, and Weber]{DBLP:conf/lrec/ArdilaBDKMHMSTW20}
Rosana Ardila, Megan Branson, Kelly Davis, Michael Kohler, Josh Meyer, Michael Henretty, Reuben Morais, Lindsay Saunders, Francis~M. Tyers, and Gregor Weber.
\newblock Common voice: {A} massively-multilingual speech corpus.
\newblock In Nicoletta Calzolari, Fr{\'{e}}d{\'{e}}ric B{\'{e}}chet, Philippe Blache, Khalid Choukri, Christopher Cieri, Thierry Declerck, Sara Goggi, Hitoshi Isahara, Bente Maegaard, Joseph Mariani, H{\'{e}}l{\`{e}}ne Mazo, Asunci{\'{o}}n Moreno, Jan Odijk, and Stelios Piperidis (eds.), \emph{Proceedings of The 12th Language Resources and Evaluation Conference, {LREC} 2020, Marseille, France, May 11-16, 2020}, pp.\  4218--4222. European Language Resources Association, 2020.
\newblock URL \url{https://aclanthology.org/2020.lrec-1.520/}.

\bibitem[Bai et~al.(2025)Bai, Cai, Chen, Chen, Chen, Cheng, Deng, Ding, Gao, Ge, Ge, Guo, Huang, Huang, Huang, Hui, Jiang, Li, Li, Li, Li, Lin, Lin, Liu, Liu, Liu, Liu, Liu, Liu, Lu, Luo, Lv, Men, Meng, Ren, Ren, Song, Sun, Tang, Tu, Wan, Wang, Wang, Wang, Wang, Xie, Xu, Xu, Xu, Yang, Yang, Yang, Yang, Yu, Zhang, Zhang, Zhang, Zheng, Zhong, Zhou, Zhou, Zhou, Zhu, and Zhu]{bai2025qwen3vltechnicalreport}
Shuai Bai, Yuxuan Cai, Ruizhe Chen, Keqin Chen, Xionghui Chen, Zesen Cheng, Lianghao Deng, Wei Ding, Chang Gao, Chunjiang Ge, Wenbin Ge, Zhifang Guo, Qidong Huang, Jie Huang, Fei Huang, Binyuan Hui, Shutong Jiang, Zhaohai Li, Mingsheng Li, Mei Li, Kaixin Li, Zicheng Lin, Junyang Lin, Xuejing Liu, Jiawei Liu, Chenglong Liu, Yang Liu, Dayiheng Liu, Shixuan Liu, Dunjie Lu, Ruilin Luo, Chenxu Lv, Rui Men, Lingchen Meng, Xuancheng Ren, Xingzhang Ren, Sibo Song, Yuchong Sun, Jun Tang, Jianhong Tu, Jianqiang Wan, Peng Wang, Pengfei Wang, Qiuyue Wang, Yuxuan Wang, Tianbao Xie, Yiheng Xu, Haiyang Xu, Jin Xu, Zhibo Yang, Mingkun Yang, Jianxin Yang, An~Yang, Bowen Yu, Fei Zhang, Hang Zhang, Xi~Zhang, Bo~Zheng, Humen Zhong, Jingren Zhou, Fan Zhou, Jing Zhou, Yuanzhi Zhu, and Ke~Zhu.
\newblock Qwen3-vl technical report, 2025.
\newblock URL \url{https://arxiv.org/abs/2511.21631}.

\bibitem[Borsos et~al.(2023)Borsos, Marinier, Vincent, Kharitonov, Pietquin, Sharifi, Roblek, Teboul, Grangier, Tagliasacchi, and Zeghidour]{DBLP:journals/taslp/BorsosMVKPSRTGTZ23}
Zal{\'{a}}n Borsos, Rapha{\"{e}}l Marinier, Damien Vincent, Eugene Kharitonov, Olivier Pietquin, Matthew Sharifi, Dominik Roblek, Olivier Teboul, David Grangier, Marco Tagliasacchi, and Neil Zeghidour.
\newblock Audiolm: {A} language modeling approach to audio generation.
\newblock \emph{{IEEE} {ACM} Trans. Audio Speech Lang. Process.}, 31:\penalty0 2523--2533, 2023.
\newblock \doi{10.1109/TASLP.2023.3288409}.
\newblock URL \url{https://doi.org/10.1109/TASLP.2023.3288409}.

\bibitem[Chen et~al.(2025)Chen, Hao, Liu, Huang, Zeng, Yu, Li, Wang, Gan, Huang, Liu, Wang, Lian, Yin, Wang, and Liu]{DBLP:journals/corr/abs-2501-12851}
Chen Chen, Xinlong Hao, Weiwen Liu, Xu~Huang, Xingshan Zeng, Shuai Yu, Dexun Li, Shuai Wang, Weinan Gan, Yuefeng Huang, Wulong Liu, Xinzhi Wang, Defu Lian, Baoqun Yin, Yasheng Wang, and Wu~Liu.
\newblock Acebench: Who wins the match point in tool learning?
\newblock \emph{CoRR}, abs/2501.12851, 2025.
\newblock \doi{10.48550/ARXIV.2501.12851}.
\newblock URL \url{https://doi.org/10.48550/arXiv.2501.12851}.

\bibitem[Chen et~al.(2024{\natexlab{a}})Chen, Ma, Yan, Liang, Li, Xu, Niu, Zhu, Yang, Liu, et~al.]{chen2024slam}
Wenxi Chen, Ziyang Ma, Ruiqi Yan, Yuzhe Liang, Xiquan Li, Ruiyang Xu, Zhikang Niu, Yanqiao Zhu, Yifan Yang, Zhanxun Liu, et~al.
\newblock Slam-omni: Timbre-controllable voice interaction system with single-stage training.
\newblock \emph{arXiv preprint arXiv:2412.15649}, 2024{\natexlab{a}}.

\bibitem[Chen et~al.(2024{\natexlab{b}})Chen, Yue, Zhang, Gao, Tan, and Li]{DBLP:journals/corr/abs-2410-17196}
Yiming Chen, Xianghu Yue, Chen Zhang, Xiaoxue Gao, Robby~T. Tan, and Haizhou Li.
\newblock Voicebench: Benchmarking llm-based voice assistants.
\newblock \emph{CoRR}, abs/2410.17196, 2024{\natexlab{b}}.
\newblock \doi{10.48550/ARXIV.2410.17196}.
\newblock URL \url{https://doi.org/10.48550/arXiv.2410.17196}.

\bibitem[D{\'{e}}fossez et~al.(2024)D{\'{e}}fossez, Mazar{\'{e}}, Orsini, Royer, P{\'{e}}rez, J{\'{e}}gou, Grave, and Zeghidour]{DBLP:journals/corr/abs-2410-00037}
Alexandre D{\'{e}}fossez, Laurent Mazar{\'{e}}, Manu Orsini, Am{\'{e}}lie Royer, Patrick P{\'{e}}rez, Herv{\'{e}} J{\'{e}}gou, Edouard Grave, and Neil Zeghidour.
\newblock Moshi: a speech-text foundation model for real-time dialogue.
\newblock \emph{CoRR}, abs/2410.00037, 2024.
\newblock \doi{10.48550/ARXIV.2410.00037}.
\newblock URL \url{https://doi.org/10.48550/arXiv.2410.00037}.

\bibitem[D\'efossez et~al.(2024)D\'efossez, Mazar\'e, Orsini, Royer, P\'erez, J\'egou, Grave, and Zeghidour]{kyutai2024moshi}
Alexandre D\'efossez, Laurent Mazar\'e, Manu Orsini, Am\'elie Royer, Patrick P\'erez, Herv\'e J\'egou, Edouard Grave, and Neil Zeghidour.
\newblock Moshi: a speech-text foundation model for real-time dialogue.
\newblock Technical report, 2024.
\newblock URL \url{https://arxiv.org/abs/2410.00037}.

\bibitem[Du et~al.(2024{\natexlab{a}})Du, Chen, Zhang, Hu, Lu, Yang, Hu, Zheng, Gu, Ma, Gao, and Yan]{DBLP:journals/corr/abs-2407-05407}
Zhihao Du, Qian Chen, Shiliang Zhang, Kai Hu, Heng Lu, Yexin Yang, Hangrui Hu, Siqi Zheng, Yue Gu, Ziyang Ma, Zhifu Gao, and Zhijie Yan.
\newblock Cosyvoice: {A} scalable multilingual zero-shot text-to-speech synthesizer based on supervised semantic tokens.
\newblock \emph{CoRR}, abs/2407.05407, 2024{\natexlab{a}}.
\newblock \doi{10.48550/ARXIV.2407.05407}.
\newblock URL \url{https://doi.org/10.48550/arXiv.2407.05407}.

\bibitem[Du et~al.(2024{\natexlab{b}})Du, Wang, Chen, Shi, Lv, Zhao, Gao, Yang, Gao, Wang, Yu, Liu, Sheng, Gu, Deng, Wang, Zhang, Yan, and Zhou]{DBLP:journals/corr/abs-2412-10117}
Zhihao Du, Yuxuan Wang, Qian Chen, Xian Shi, Xiang Lv, Tianyu Zhao, Zhifu Gao, Yexin Yang, Changfeng Gao, Hui Wang, Fan Yu, Huadai Liu, Zhengyan Sheng, Yue Gu, Chong Deng, Wen Wang, Shiliang Zhang, Zhijie Yan, and Jingren Zhou.
\newblock Cosyvoice 2: Scalable streaming speech synthesis with large language models.
\newblock \emph{CoRR}, abs/2412.10117, 2024{\natexlab{b}}.
\newblock \doi{10.48550/ARXIV.2412.10117}.
\newblock URL \url{https://doi.org/10.48550/arXiv.2412.10117}.

\bibitem[Du et~al.(2025)Du, Gao, Wang, Yu, Zhao, Wang, Lv, Wang, Ni, Shi, An, Yang, Li, Chen, Gao, Chen, Gu, Chen, Chen, Zhang, Wang, and Ye]{DBLP:journals/corr/abs-2505-17589}
Zhihao Du, Changfeng Gao, Yuxuan Wang, Fan Yu, Tianyu Zhao, Hao Wang, Xiang Lv, Hui Wang, Chongjia Ni, Xian Shi, Keyu An, Guanrou Yang, Yabin Li, Yanni Chen, Zhifu Gao, Qian Chen, Yue Gu, Mengzhe Chen, Yafeng Chen, Shiliang Zhang, Wen Wang, and Jieping Ye.
\newblock Cosyvoice 3: Towards in-the-wild speech generation via scaling-up and post-training.
\newblock \emph{CoRR}, abs/2505.17589, 2025.
\newblock \doi{10.48550/ARXIV.2505.17589}.
\newblock URL \url{https://doi.org/10.48550/arXiv.2505.17589}.

\bibitem[Goel et~al.(2025)Goel, Ghosh, Kim, Kumar, Kong, Lee, Yang, Duraiswami, Manocha, Valle, and Catanzaro]{DBLP:journals/corr/abs-2507-08128}
Arushi Goel, Sreyan Ghosh, Jaehyeon Kim, Sonal Kumar, Zhifeng Kong, Sang{-}gil Lee, Chao{-}Han~Huck Yang, Ramani Duraiswami, Dinesh Manocha, Rafael Valle, and Bryan Catanzaro.
\newblock Audio flamingo 3: Advancing audio intelligence with fully open large audio language models.
\newblock \emph{CoRR}, abs/2507.08128, 2025.
\newblock \doi{10.48550/ARXIV.2507.08128}.
\newblock URL \url{https://doi.org/10.48550/arXiv.2507.08128}.

\bibitem[KimiTeam et~al.(2025)KimiTeam, Ding, Ju, Leng, Liu, Liu, Shang, Shen, Song, Tan, Tang, Wang, Wei, Xin, Xu, Yu, Zhang, Zhou, Charles, Chen, Chen, Du, He, Hu, Lai, Li, Liu, Sun, Wang, Wang, Wu, Wu, Yang, Yang, Yang, Yang, Yin, Yuan, Zhang, and Zhou]{kimiteam2025kimiaudiotechnicalreport}
KimiTeam, Ding Ding, Zeqian Ju, Yichong Leng, Songxiang Liu, Tong Liu, Zeyu Shang, Kai Shen, Wei Song, Xu~Tan, Heyi Tang, Zhengtao Wang, Chu Wei, Yifei Xin, Xinran Xu, Jianwei Yu, Yutao Zhang, Xinyu Zhou, Y.~Charles, Jun Chen, Yanru Chen, Yulun Du, Weiran He, Zhenxing Hu, Guokun Lai, Qingcheng Li, Yangyang Liu, Weidong Sun, Jianzhou Wang, Yuzhi Wang, Yuefeng Wu, Yuxin Wu, Dongchao Yang, Hao Yang, Ying Yang, Zhilin Yang, Aoxiong Yin, Ruibin Yuan, Yutong Zhang, and Zaida Zhou.
\newblock Kimi-audio technical report, 2025.
\newblock URL \url{https://arxiv.org/abs/2504.18425}.

\bibitem[Kong et~al.(2020)Kong, Kim, and Bae]{DBLP:conf/nips/KongKB20}
Jungil Kong, Jaehyeon Kim, and Jaekyoung Bae.
\newblock Hifi-gan: Generative adversarial networks for efficient and high fidelity speech synthesis.
\newblock In Hugo Larochelle, Marc'Aurelio Ranzato, Raia Hadsell, Maria{-}Florina Balcan, and Hsuan{-}Tien Lin (eds.), \emph{Advances in Neural Information Processing Systems 33: Annual Conference on Neural Information Processing Systems 2020, NeurIPS 2020, December 6-12, 2020, virtual}, 2020.
\newblock URL \url{https://proceedings.neurips.cc/paper/2020/hash/c5d736809766d46260d816d8dbc9eb44-Abstract.html}.

\bibitem[Kumar et~al.(2025)Kumar, Sedl{\'{a}}cek, Lokegaonkar, L{\'{o}}pez, Yu, Anand, Ryu, Chen, Plicka, Hlav{\'{a}}cek, Ellingwood, Udupa, Hou, Ferner, Barahona, Bola{\~{n}}os, Rahi, Herrera{-}Alarc{\'{o}}n, Dixit, Patil, Deshmukh, Koroshinadze, Liu, Perera, Zanou, Stafylakis, Chung, Harwath, Zhang, Manocha, Lozano{-}Diez, Kesiraju, Ghosh, and Duraiswami]{DBLP:journals/corr/abs-2508-13992}
Sonal Kumar, Simon Sedl{\'{a}}cek, Vaibhavi Lokegaonkar, Fernando L{\'{o}}pez, Wenyi Yu, Nishit Anand, Hyeonggon Ryu, Lichang Chen, Maxim Plicka, Miroslav Hlav{\'{a}}cek, William~Fineas Ellingwood, Sathvik Udupa, Siyuan Hou, Allison Ferner, Sara Barahona, Cecilia Bola{\~{n}}os, Satish Rahi, Laura Herrera{-}Alarc{\'{o}}n, Satvik Dixit, Rupali~S. Patil, Soham Deshmukh, Lasha Koroshinadze, Yao Liu, Leibny Paola~Garcia Perera, Eleni Zanou, Themos Stafylakis, Joon~Son Chung, David Harwath, Chao Zhang, Dinesh Manocha, Alicia Lozano{-}Diez, Santosh Kesiraju, Sreyan Ghosh, and Ramani Duraiswami.
\newblock Mmau-pro: {A} challenging and comprehensive benchmark for holistic evaluation of audio general intelligence.
\newblock \emph{CoRR}, abs/2508.13992, 2025.
\newblock \doi{10.48550/ARXIV.2508.13992}.
\newblock URL \url{https://doi.org/10.48550/arXiv.2508.13992}.

\bibitem[Li et~al.(2025{\natexlab{a}})Li, Liu, Zhang, Fang, Pan, Wang, Liang, Li, Lin, Dong, Xu, Sun, Zhou, and Chen]{DBLP:journals/corr/abs-2502-17239}
Tianpeng Li, Jun Liu, Tao Zhang, Yuanbo Fang, Da~Pan, Mingrui Wang, Zheng Liang, Zehuan Li, Mingan Lin, Guosheng Dong, Jianhua Xu, Haoze Sun, Zenan Zhou, and Weipeng Chen.
\newblock Baichuan-audio: {A} unified framework for end-to-end speech interaction.
\newblock \emph{CoRR}, abs/2502.17239, 2025{\natexlab{a}}.
\newblock \doi{10.48550/ARXIV.2502.17239}.
\newblock URL \url{https://doi.org/10.48550/arXiv.2502.17239}.

\bibitem[Li et~al.(2023)Li, Zhang, Dubois, Taori, Gulrajani, Guestrin, Liang, and Hashimoto]{alpaca_eval}
Xuechen Li, Tianyi Zhang, Yann Dubois, Rohan Taori, Ishaan Gulrajani, Carlos Guestrin, Percy Liang, and Tatsunori~B. Hashimoto.
\newblock Alpacaeval: An automatic evaluator of instruction-following models.
\newblock \url{https://github.com/tatsu-lab/alpaca_eval}, 5 2023.

\bibitem[Li et~al.(2025{\natexlab{b}})Li, Liu, Zhang, Chen, Li, Li, Liu, Ming, Dong, Pan, et~al.]{li2025baichuan}
Yadong Li, Jun Liu, Tao Zhang, Song Chen, Tianpeng Li, Zehuan Li, Lijun Liu, Lingfeng Ming, Guosheng Dong, Da~Pan, et~al.
\newblock Baichuan-omni-1.5 technical report.
\newblock \emph{arXiv preprint arXiv:2501.15368}, 2025{\natexlab{b}}.

\bibitem[Lipman et~al.(2023)Lipman, Chen, Ben{-}Hamu, Nickel, and Le]{DBLP:conf/iclr/LipmanCBNL23}
Yaron Lipman, Ricky T.~Q. Chen, Heli Ben{-}Hamu, Maximilian Nickel, and Matthew Le.
\newblock Flow matching for generative modeling.
\newblock In \emph{The Eleventh International Conference on Learning Representations, {ICLR} 2023, Kigali, Rwanda, May 1-5, 2023}. OpenReview.net, 2023.
\newblock URL \url{https://openreview.net/forum?id=PqvMRDCJT9t}.

\bibitem[Liu et~al.(2023)Liu, Iter, Xu, Wang, Xu, and Zhu]{DBLP:conf/emnlp/LiuIXWXZ23}
Yang Liu, Dan Iter, Yichong Xu, Shuohang Wang, Ruochen Xu, and Chenguang Zhu.
\newblock G-eval: {NLG} evaluation using gpt-4 with better human alignment.
\newblock In Houda Bouamor, Juan Pino, and Kalika Bali (eds.), \emph{Proceedings of the 2023 Conference on Empirical Methods in Natural Language Processing, {EMNLP} 2023, Singapore, December 6-10, 2023}, pp.\  2511--2522. Association for Computational Linguistics, 2023.
\newblock \doi{10.18653/V1/2023.EMNLP-MAIN.153}.
\newblock URL \url{https://doi.org/10.18653/v1/2023.emnlp-main.153}.

\bibitem[{OpenAI}(2024b)]{openai2024hello}
{OpenAI}.
\newblock {Hello GPT-4o}, 2024b.
\newblock URL \url{https://openai.com/index/hello-gpt-4o/}.

\bibitem[Panayotov et~al.(2015)Panayotov, Chen, Povey, and Khudanpur]{DBLP:conf/icassp/PanayotovCPK15}
Vassil Panayotov, Guoguo Chen, Daniel Povey, and Sanjeev Khudanpur.
\newblock Librispeech: An {ASR} corpus based on public domain audio books.
\newblock In \emph{2015 {IEEE} International Conference on Acoustics, Speech and Signal Processing, {ICASSP} 2015, South Brisbane, Queensland, Australia, April 19-24, 2015}, pp.\  5206--5210. {IEEE}, 2015.
\newblock \doi{10.1109/ICASSP.2015.7178964}.
\newblock URL \url{https://doi.org/10.1109/ICASSP.2015.7178964}.

\bibitem[Patil et~al.(2025)Patil, Mao, Yan, Ji, Suresh, Stoica, and Gonzalez]{patil2025bfcl}
Shishir~G. Patil, Huanzhi Mao, Fanjia Yan, Charlie~Cheng{-}Jie Ji, Vishnu Suresh, Ion Stoica, and Joseph~E. Gonzalez.
\newblock The berkeley function calling leaderboard {(BFCL):} from tool use to agentic evaluation of large language models.
\newblock In \emph{Forty-second International Conference on Machine Learning, {ICML} 2025, Vancouver, BC, Canada, July 13-19, 2025}. OpenReview.net, 2025.
\newblock URL \url{https://openreview.net/forum?id=2GmDdhBdDk}.

\bibitem[Radford et~al.(2022)Radford, Kim, Xu, Brockman, McLeavey, and Sutskever]{radford2022whisper}
Alec Radford, Jong~Wook Kim, Tao Xu, Greg Brockman, Christine McLeavey, and Ilya Sutskever.
\newblock Robust speech recognition via large-scale weak supervision, 2022.
\newblock URL \url{https://arxiv.org/abs/2212.04356}.

\bibitem[Rafailov et~al.(2023)Rafailov, Sharma, Mitchell, Manning, Ermon, and Finn]{DBLP:conf/nips/RafailovSMMEF23}
Rafael Rafailov, Archit Sharma, Eric Mitchell, Christopher~D. Manning, Stefano Ermon, and Chelsea Finn.
\newblock Direct preference optimization: Your language model is secretly a reward model.
\newblock In Alice Oh, Tristan Naumann, Amir Globerson, Kate Saenko, Moritz Hardt, and Sergey Levine (eds.), \emph{Advances in Neural Information Processing Systems 36: Annual Conference on Neural Information Processing Systems 2023, NeurIPS 2023, New Orleans, LA, USA, December 10 - 16, 2023}, 2023.
\newblock URL \url{http://papers.nips.cc/paper\_files/paper/2023/hash/a85b405ed65c6477a4fe8302b5e06ce7-Abstract-Conference.html}.

\bibitem[Saeki et~al.(2022)Saeki, Xin, Nakata, Koriyama, Takamichi, and Saruwatari]{DBLP:conf/interspeech/SaekiXNKTS22}
Takaaki Saeki, Detai Xin, Wataru Nakata, Tomoki Koriyama, Shinnosuke Takamichi, and Hiroshi Saruwatari.
\newblock {UTMOS:} utokyo-sarulab system for voicemos challenge 2022.
\newblock In Hanseok Ko and John H.~L. Hansen (eds.), \emph{23rd Annual Conference of the International Speech Communication Association, Interspeech 2022, Incheon, Korea, September 18-22, 2022}, pp.\  4521--4525. {ISCA}, 2022.
\newblock \doi{10.21437/INTERSPEECH.2022-439}.
\newblock URL \url{https://doi.org/10.21437/Interspeech.2022-439}.

\bibitem[Sakshi et~al.(2025)Sakshi, Tyagi, Kumar, Seth, Selvakumar, Nieto, Duraiswami, Ghosh, and Manocha]{DBLP:conf/iclr/SakshiTKSSNDGM25}
S.~Sakshi, Utkarsh Tyagi, Sonal Kumar, Ashish Seth, Ramaneswaran Selvakumar, Oriol Nieto, Ramani Duraiswami, Sreyan Ghosh, and Dinesh Manocha.
\newblock {MMAU:} {A} massive multi-task audio understanding and reasoning benchmark.
\newblock In \emph{The Thirteenth International Conference on Learning Representations, {ICLR} 2025, Singapore, April 24-28, 2025}. OpenReview.net, 2025.
\newblock URL \url{https://openreview.net/forum?id=TeVAZXr3yv}.

\bibitem[Tan et~al.(2025)Tan, Chen, Wang, Deng, Zhang, Cheng, Yu, Zhang, Lv, Zhao, Zhang, Ma, Chen, Wang, Liu, Li, and Ye]{tan2025drvoiceparallelspeechtextvoice}
Chao-Hong Tan, Qian Chen, Wen Wang, Chong Deng, Qinglin Zhang, Luyao Cheng, Hai Yu, Xin Zhang, Xiang Lv, Tianyu Zhao, Chong Zhang, Yukun Ma, Yafeng Chen, Hui Wang, Jiaqing Liu, Xiangang Li, and Jieping Ye.
\newblock Drvoice: Parallel speech-text voice conversation model via dual-resolution speech representations, 2025.
\newblock URL \url{https://arxiv.org/abs/2506.09349}.

\bibitem[Team(2025{\natexlab{a}})]{DBLP:journals/corr/abs-2507-06261}
Gemini Team.
\newblock Gemini 2.5: Pushing the frontier with advanced reasoning, multimodality, long context, and next generation agentic capabilities.
\newblock \emph{CoRR}, abs/2507.06261, 2025{\natexlab{a}}.
\newblock \doi{10.48550/ARXIV.2507.06261}.
\newblock URL \url{https://doi.org/10.48550/arXiv.2507.06261}.

\bibitem[Team(2025{\natexlab{b}})]{DBLP:journals/corr/abs-2511-00279}
Meituan~LongCat Team.
\newblock Longcat-flash-omni technical report.
\newblock \emph{CoRR}, abs/2511.00279, 2025{\natexlab{b}}.

\bibitem[Wang et~al.(2025{\natexlab{a}})Wang, Wu, Li, Yang, Chen, Zhang, and Meng]{DBLP:journals/corr/abs-2506-04779}
Dingdong Wang, Jincenzi Wu, Junan Li, Dongchao Yang, Xueyuan Chen, Tianhua Zhang, and Helen Meng.
\newblock {MMSU:} {A} massive multi-task spoken language understanding and reasoning benchmark.
\newblock \emph{CoRR}, abs/2506.04779, 2025{\natexlab{a}}.
\newblock \doi{10.48550/ARXIV.2506.04779}.
\newblock URL \url{https://doi.org/10.48550/arXiv.2506.04779}.

\bibitem[Wang et~al.(2025{\natexlab{b}})Wang, Li, Fu, Zhang, Shen, Xie, Li, Sun, and Ma]{DBLP:conf/icml/WangLFZS000M25}
Xiong Wang, Yangze Li, Chaoyou Fu, Yike Zhang, Yunhang Shen, Lei Xie, Ke~Li, Xing Sun, and Long Ma.
\newblock Freeze-omni: {A} smart and low latency speech-to-speech dialogue model with frozen {LLM}.
\newblock In \emph{Forty-second International Conference on Machine Learning, {ICML} 2025, Vancouver, BC, Canada, July 13-19, 2025}. OpenReview.net, 2025{\natexlab{b}}.
\newblock URL \url{https://openreview.net/forum?id=s1EImzs5Id}.

\bibitem[Wu et~al.(2025)Wu, Yan, Hu, Yi, Feng, Tian, Shen, Yu, Zhang, Li, Chen, Liu, You, Zhang, Li, Yang, Deng, Huang, Li, Zhang, You, Li, Wan, Hu, Zhen, Chen, Yuan, Zhang, Jiang, Zhou, Yang, Li, Ma, Song, Pang, Hu, Sun, An, Wang, Gao, Ji, Li, Sun, Wen, Ren, Ma, Lu, Wang, Li, Miao, Liu, Xu, Shi, Hu, Wu, Liu, Huang, Yan, Zhang, Hao, Jia, Zhou, Sun, Wu, Wu, Yang, Yang, Lin, Li, Yang, Shi, Zhou, Gu, Li, Li, Li, Wu, Han, Tan, Pang, Fan, Liu, Cao, Lu, He, Xie, Zhao, Li, Yu, Yang, Liu, Lu, Wang, Ding, Liang, Lu, Luo, Yin, Zhan, and Zhang]{DBLP:journals/corr/abs-2507-16632}
Boyong Wu, Chao Yan, Chen Hu, Cheng Yi, Chengli Feng, Fei Tian, Feiyu Shen, Gang Yu, Haoyang Zhang, Jingbei Li, Mingrui Chen, Peng Liu, Wang You, Xiangyu~Tony Zhang, Xingyuan Li, Xuerui Yang, Yayue Deng, Yechang Huang, Yuxin Li, Yuxin Zhang, Zhao You, Brian Li, Changyi Wan, Hanpeng Hu, Jiangjie Zhen, Siyu Chen, Song Yuan, Xuelin Zhang, Yimin Jiang, Yu~Zhou, Yuxiang Yang, Bingxin Li, Buyun Ma, Changhe Song, Dongqing Pang, Guoqiang Hu, Haiyang Sun, Kang An, Na~Wang, Shuli Gao, Wei Ji, Wen Li, Wen Sun, Xuan Wen, Yong Ren, Yuankai Ma, Yufan Lu, Bin Wang, Bo~Li, Changxin Miao, Che Liu, Chen Xu, Dapeng Shi, Dingyuan Hu, Donghang Wu, Enle Liu, Guanzhe Huang, Gulin Yan, Han Zhang, Nie Hao, Haonan Jia, Hongyu Zhou, Jianjian Sun, Jiaoren Wu, Jie Wu, Jie Yang, Jin Yang, Junzhe Lin, Kaixiang Li, Lei Yang, Liying Shi, Li~Zhou, Longlong Gu, Ming Li, Mingliang Li, Mingxiao Li, Nan Wu, Qi~Han, Qinyuan Tan, Shaoliang Pang, Shengjie Fan, Siqi Liu, Tiancheng Cao, Wanying Lu, Wenqing He, Wuxun Xie, Xu~Zhao, Xueqi Li, Yanbo Yu,
  Yang Yang, Yi~Liu, Yifan Lu, Yilei Wang, Yuanhao Ding, Yuanwei Liang, Yuanwei Lu, Yuchu Luo, Yuhe Yin, Yumeng Zhan, and Yuxiang Zhang.
\newblock Step-audio 2 technical report.
\newblock \emph{CoRR}, abs/2507.16632, 2025.
\newblock \doi{10.48550/ARXIV.2507.16632}.
\newblock URL \url{https://doi.org/10.48550/arXiv.2507.16632}.

\bibitem[Xiao et~al.(2024)Xiao, Liu, Zhang, and Xing]{DBLP:conf/acl/XiaoLZX24}
Shitao Xiao, Zheng Liu, Peitian Zhang, and Xingrun Xing.
\newblock Lm-cocktail: Resilient tuning of language models via model merging.
\newblock In Lun{-}Wei Ku, Andre Martins, and Vivek Srikumar (eds.), \emph{Findings of the Association for Computational Linguistics, {ACL} 2024, Bangkok, Thailand and virtual meeting, August 11-16, 2024}, pp.\  2474--2488. Association for Computational Linguistics, 2024.
\newblock \doi{10.18653/V1/2024.FINDINGS-ACL.145}.
\newblock URL \url{https://doi.org/10.18653/v1/2024.findings-acl.145}.

\bibitem[Xiaomi(2025)]{coreteam2025mimoaudio}
LLM-Core-Team Xiaomi.
\newblock Mimo-audio: Audio language models are few-shot learners, 2025.
\newblock URL \url{https://github.com/XiaomiMiMo/MiMo-Audio}.

\bibitem[Xu et~al.(2025)Xu, Guo, He, Hu, He, Bai, Chen, Wang, Fan, Dang, Zhang, Wang, Chu, and Lin]{DBLP:journals/corr/abs-2503-20215}
Jin Xu, Zhifang Guo, Jinzheng He, Hangrui Hu, Ting He, Shuai Bai, Keqin Chen, Jialin Wang, Yang Fan, Kai Dang, Bin Zhang, Xiong Wang, Yunfei Chu, and Junyang Lin.
\newblock Qwen2.5-omni technical report.
\newblock \emph{CoRR}, abs/2503.20215, 2025.
\newblock \doi{10.48550/ARXIV.2503.20215}.
\newblock URL \url{https://doi.org/10.48550/arXiv.2503.20215}.

\bibitem[Yang et~al.(2025)Yang, Li, Yang, Zhang, Hui, Zheng, Yu, Gao, Huang, Lv, Zheng, Liu, Zhou, Huang, Hu, Ge, Wei, Lin, Tang, Yang, Tu, Zhang, Yang, Yang, Zhou, Lin, Dang, Bao, Yang, Yu, Deng, Li, Xue, Li, Zhang, Wang, Zhu, Men, Gao, Liu, Luo, Li, Tang, Yin, Ren, Wang, Zhang, Ren, Fan, Su, Zhang, Zhang, Wan, Liu, Wang, Cui, Zhang, Zhou, and Qiu]{DBLP:journals/corr/abs-2505-09388}
An~Yang, Anfeng Li, Baosong Yang, Beichen Zhang, Binyuan Hui, Bo~Zheng, Bowen Yu, Chang Gao, Chengen Huang, Chenxu Lv, Chujie Zheng, Dayiheng Liu, Fan Zhou, Fei Huang, Feng Hu, Hao Ge, Haoran Wei, Huan Lin, Jialong Tang, Jian Yang, Jianhong Tu, Jianwei Zhang, Jian Yang, Jiaxi Yang, Jingren Zhou, Junyang Lin, Kai Dang, Keqin Bao, Kexin Yang, Le~Yu, Lianghao Deng, Mei Li, Mingfeng Xue, Mingze Li, Pei Zhang, Peng Wang, Qin Zhu, Rui Men, Ruize Gao, Shixuan Liu, Shuang Luo, Tianhao Li, Tianyi Tang, Wenbiao Yin, Xingzhang Ren, Xinyu Wang, Xinyu Zhang, Xuancheng Ren, Yang Fan, Yang Su, Yichang Zhang, Yinger Zhang, Yu~Wan, Yuqiong Liu, Zekun Wang, Zeyu Cui, Zhenru Zhang, Zhipeng Zhou, and Zihan Qiu.
\newblock Qwen3 technical report.
\newblock \emph{CoRR}, abs/2505.09388, 2025.
\newblock \doi{10.48550/ARXIV.2505.09388}.
\newblock URL \url{https://doi.org/10.48550/arXiv.2505.09388}.

\bibitem[Yao et~al.(2024)Yao, Yu, Zhang, Wang, Cui, Zhu, Cai, Li, Zhao, He, et~al.]{yao2024minicpm}
Yuan Yao, Tianyu Yu, Ao~Zhang, Chongyi Wang, Junbo Cui, Hongji Zhu, Tianchi Cai, Haoyu Li, Weilin Zhao, Zhihui He, et~al.
\newblock Minicpm-v: A gpt-4v level mllm on your phone.
\newblock \emph{arXiv preprint arXiv:2408.01800}, 2024.

\bibitem[Zeng et~al.(2024)Zeng, Du, Liu, Wang, Jiang, Zhao, Dong, and Tang]{DBLP:journals/corr/abs-2412-02612}
Aohan Zeng, Zhengxiao Du, Mingdao Liu, Kedong Wang, Shengmin Jiang, Lei Zhao, Yuxiao Dong, and Jie Tang.
\newblock Glm-4-voice: Towards intelligent and human-like end-to-end spoken chatbot.
\newblock \emph{CoRR}, abs/2412.02612, 2024.
\newblock \doi{10.48550/ARXIV.2412.02612}.
\newblock URL \url{https://doi.org/10.48550/arXiv.2412.02612}.

\bibitem[Zhan et~al.(2025)Zhan, Han, Xie, Wang, Zhang, Huang, Shi, Wang, Song, Cheng, Li, Song, Qiu, and Zheng]{DBLP:journals/corr/abs-2509-09716}
Jun Zhan, Mingyang Han, Yuxuan Xie, Chen Wang, Dong Zhang, Kexin Huang, Haoxiang Shi, DongXiao Wang, Tengtao Song, Qinyuan Cheng, Shimin Li, Jun Song, Xipeng Qiu, and Bo~Zheng.
\newblock Vstyle: {A} benchmark for voice style adaptation with spoken instructions.
\newblock \emph{CoRR}, abs/2509.09716, 2025.
\newblock \doi{10.48550/ARXIV.2509.09716}.
\newblock URL \url{https://doi.org/10.48550/arXiv.2509.09716}.

\bibitem[Zhang et~al.(2023{\natexlab{a}})Zhang, Li, Zhang, Zhan, Wang, Zhou, and Qiu]{DBLP:conf/emnlp/ZhangLZZWZQ23}
Dong Zhang, Shimin Li, Xin Zhang, Jun Zhan, Pengyu Wang, Yaqian Zhou, and Xipeng Qiu.
\newblock Speechgpt: Empowering large language models with intrinsic cross-modal conversational abilities.
\newblock In Houda Bouamor, Juan Pino, and Kalika Bali (eds.), \emph{Findings of the Association for Computational Linguistics: {EMNLP} 2023, Singapore, December 6-10, 2023}, pp.\  15757--15773. Association for Computational Linguistics, 2023{\natexlab{a}}.
\newblock \doi{10.18653/V1/2023.FINDINGS-EMNLP.1055}.
\newblock URL \url{https://doi.org/10.18653/v1/2023.findings-emnlp.1055}.

\bibitem[Zhang et~al.(2025)Zhang, Cheng, Deng, Chen, Wang, Zheng, Liu, Yu, Tan, Du, and Zhang]{zhang-etal-2025-omniflatten}
Qinglin Zhang, Luyao Cheng, Chong Deng, Qian Chen, Wen Wang, Siqi Zheng, Jiaqing Liu, Hai Yu, Chao-Hong Tan, Zhihao Du, and ShiLiang Zhang.
\newblock {O}mni{F}latten: An end-to-end {GPT} model for seamless voice conversation.
\newblock In Wanxiang Che, Joyce Nabende, Ekaterina Shutova, and Mohammad~Taher Pilehvar (eds.), \emph{Proceedings of the 63rd Annual Meeting of the Association for Computational Linguistics (Volume 1: Long Papers)}, pp.\  14570--14580, Vienna, Austria, July 2025. Association for Computational Linguistics.
\newblock ISBN 979-8-89176-251-0.
\newblock \doi{10.18653/v1/2025.acl-long.709}.
\newblock URL \url{https://aclanthology.org/2025.acl-long.709/}.

\bibitem[Zhang et~al.(2023{\natexlab{b}})Zhang, Zhang, Li, Zhou, and Qiu]{DBLP:journals/corr/abs-2308-16692}
Xin Zhang, Dong Zhang, Shimin Li, Yaqian Zhou, and Xipeng Qiu.
\newblock Speechtokenizer: Unified speech tokenizer for speech large language models.
\newblock \emph{CoRR}, abs/2308.16692, 2023{\natexlab{b}}.
\newblock \doi{10.48550/ARXIV.2308.16692}.
\newblock URL \url{https://doi.org/10.48550/arXiv.2308.16692}.

\end{thebibliography}
\bibliographystyle{colm2024_conference}

\end{document}